\documentclass[10pt,twocolumn,letterpaper]{article}
\usepackage[normalem]{ulem}
\usepackage{tabularray}
\usepackage{overpic}
\usepackage{multirow}
\usepackage{colortbl}
\usepackage{authblk}
\usepackage[accsupp]{axessibility}

\usepackage[pagenumbers]{iccv} % To force page numbers, e.g. for an arXiv version

\definecolor{iccvblue}{rgb}{0.21,0.49,0.74}
\usepackage[pagebackref,breaklinks,colorlinks,allcolors=iccvblue]{hyperref}

\definecolor{colorfirst}{rgb}{.922,.835,.820} % green
\definecolor{colorsecond}{rgb}{.886,.941,.851} % yellow
\definecolor{colorthird}{rgb}{0.76, 0.87, 0.92} % white

\newcommand{\cellfirst}{\cellcolor{colorfirst}}
\newcommand{\cellsecond}{\cellcolor{colorsecond}}

\newcommand{\textfirst}{\colorbox{colorfirst}}
\newcommand{\secondtext}{\colorbox{colorsecond}}

\def\paperID{10743} % *** Enter the Paper ID here
\def\confName{ICCV}
\def\confYear{2025}

\title{G-DexGrasp: Generalizable Dexterous Grasping Synthesis Via Part-Aware Prior Retrieval and Prior-Assisted Generation}

\author{Juntao Jian$^{1,2\star}$, \quad Xiuping Liu$^{2\star}$,  \quad Zixuan Chen$^{2\star}$, \quad Manyi Li$^3${$^{\dagger}$}, \quad Jian Liu$^4${$^{\dagger}$}, \quad Ruizhen Hu$^1$\\
$^1$Shenzhen University \qquad\qquad $^2$Dalian University of Technology\\
$^3$Shandong University \qquad\qquad $^4$Shenyang University of Technology\\
}

\newcommand\blfootnote[1]{%
	\begingroup
	\renewcommand\thefootnote{}\footnote{#1}
	\addtocounter{footnote}{-1}
	\endgroup
}

\begin{document}
\maketitle
\begin{abstract}
Recent advances in dexterous grasping synthesis have demonstrated significant progress in producing reasonable and plausible grasps for many task purposes. But it remains challenging to generalize to unseen object categories and diverse task instructions. In this paper, we propose G-DexGrasp, a retrieval-augmented generation approach that can produce high-quality dexterous hand configurations for unseen object categories and language-based task instructions. The key is to retrieve generalizable grasping priors, including the fine-grained contact part and the affordance-related distribution of relevant grasping instances, for the following synthesis pipeline. Specifically, the fine-grained contact part and affordance act as generalizable guidance to infer reasonable grasping configurations for unseen objects with a generative model, while the relevant grasping distribution plays as regularization to guarantee the plausibility of synthesized grasps during the subsequent refinement optimization. Our comparison experiments validate the effectiveness of our key designs for generalization and demonstrate the remarkable performance against the existing approaches. Project page: \url{https://g-dexgrasp.github.io/}
\end{abstract}

\blfootnote{$^{\star}$ \scriptsize Equal contribution}\\
\blfootnote{$^{\dagger}$ \scriptsize Corresponding Authors: manyili@sdu.edu.cn, jianliu2006@gmail.com}

\section{Introduction}
\label{sec:intro}

\begin{figure}[!t]
   \includegraphics[width=1.05\linewidth]{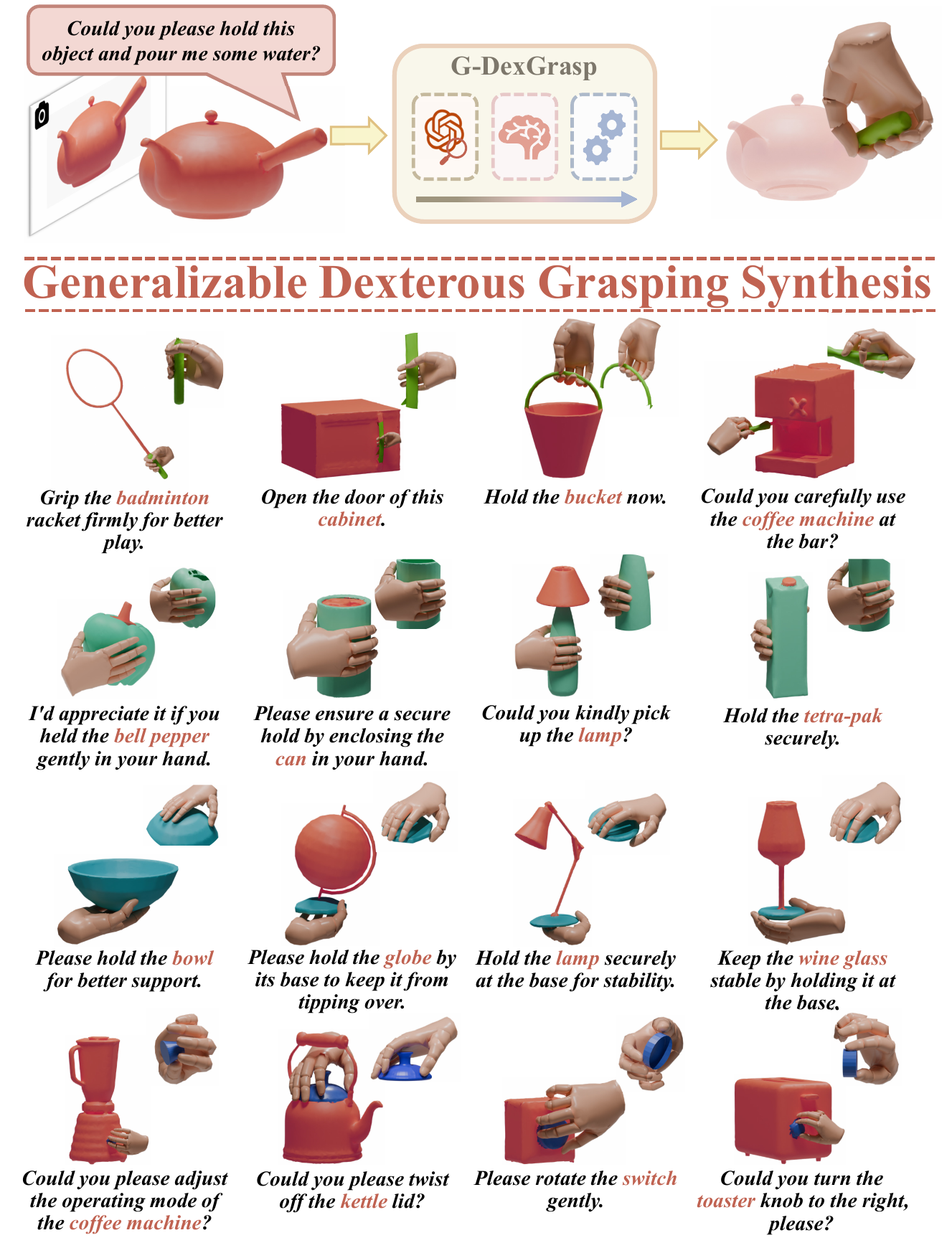}
   \caption{Given the \emph{diverse language-based task instructions and target objects of unseen categories}, \textbf{G-DexGrasp} can synthesize high-quality dexterous grasps that can stably hold the object and are semantically aligned with the specified instruction.}
   \label{fig:teaser}
   \vspace{-10pt}
\end{figure} 

Dexterous grasping synthesis ~\cite{Hasson2019Obman, Karunratanakul2020GraspingField, She2022LearningHR, Li2022GenDexGraspGD, Wang2022DexGraspNetAL, Turpin2022GraspD, Wei2022DVGGDV, Geng2023UniDexGrasp++ID, Xu2023UniDexGraspUR,Rajeswaran2017LearningCD} aims to generate reasonable and plausible solutions for the dexterous hands of robots in grasping and manipulating objects. %The high flexibility of dexterous hands supports human-like object grasping, which can effectively assist object manipulation in a wide variety of tasks. However, it also poses key challenges for dexterous grasping synthesis, since it is difficult to generate effective and reliable grasp configurations of the high-freedom dexterous hands to deal with all kinds of objects and task intents. Although 
A variety of works have proposed analysis-based ~\cite{Wang2022DexGraspNetAL, Liu2021SynthesizingDA, Dai2018SynthsisOp} or learning-based approaches ~\cite{Hampali2019HOnnotateAM, Duan2021RoboticsDG, Liu2019GeneratingGP, Chao2021DexYCBAB, Wei2022DVGGDV, Li2022EfficientGraspAU} to synthesize dexterous hand grasps. However, they often focus on producing stable and diverse grasping solutions, but ignore the functionality goal of the grasps, i.e. whether the synthesized grasps can facilitate the following manipulation tasks with all kinds of objects.

To deal with various task intents and target objects, the researchers collect large amounts of annotations and learn to guide the synthesis of dexterous hand grasping. Some works~\cite{Brahmbhatt2020ContactPoseAD, Taheri2020GRABAD, Yang2022OakInkAL, Jian2023AffordPoseAL} classify the task intents into several categories and train conditional generative models to produce dexterous hand configurations. In parallel, some recent works ~\cite{Chang2024Text2GraspGS, Li2024SemGraspSG, Wei2024DexGYGrasp} utilize the powerful pre-trained large language models to generate natural language descriptions of the detailed grasping and synthesize hand configurations corresponding to the descriptions. These recent advances have shown significant progress in dealing with a variety of task purposes, but still struggle to effectively generalize to unseen object categories for task-driven grasping synthesis.

In this paper, we propose G-DexGrasp, a \textbf{G}eneralizable retrieval-augmented \textbf{Dex}terous \textbf{Grasp}ing synthesis approach that can produce high-quality dexterous grasps for objects of unseen categories and unstructured task instructions, as shown in Figure~\ref{fig:teaser}. Considering the significant variations in object structures and shapes, the key is to retrieve \emph{generalizable grasping prior}, which encompasses the fine-grained contact part as well as the affordance-related distribution of relevant grasping instances, aiming to assist the subsequent pipeline. The fine-grained contact part and affordance act as generalizable guidance to infer reasonable grasp configuration for unseen objects with a generative model, while the relevant grasping distribution plays as regularization to guarantee the plausibility of synthesized grasps during refinement optimization.

Our approach operates in three stages. The first stage analyzes the task instruction and target object to retrieve relevant grasping instances and organize the generalizable grasping prior. The second stage trains a generative model to estimate the coarse hand configuration, conditioned on fine-grained grasping guidance. The third stage takes the coarse estimation as initialization and optimizes to refine the parameters of dexterous hands regularized by the retrieved grasping prior. It is worth noting that our approach exploits the potential of training set to facilitate generalization without requiring extra labor-intensive joint-level annotations for more objects. Comparisons against existing approaches validates the effectiveness of our retrieval-augmented dexterous grasping synthesis. The ablation study confirms the effectiveness of our key designs in achieving remarkable generalization performance.

Our contributions are summarized as follows:
\begin{itemize}
\item We present G-DexGrasp, a retrieval-augmented generation approach that can produce high-quality dexterous grasps for unseen object categories and task instructions.
\item We propose generalizable grasping prior, which acts as fine-grained guidance and semantic-aware regularization for the generalization of dexterous grasping synthesis.
\item We provide extensive experiments to validate the effectiveness of our key designs in achieving remarkable performance against existing approaches.
\end{itemize}
\section{Related Work}
\label{sec:related_work}

%-------------------------------------------------------------------------
\subsection{Task-Agnostic Dexterous Grasping}

% Manyi' summary
% (1) Some use optimization-based methods.
% (2) Some use learning-based methods. Among them, some focus on developing different networks, i.e. GAN, VAE, Diffusion, while some design intermediate representations, i.e. contact map, touch code.

% \noindent\textbf{Optimization-based Grasping}
Due to the high flexibility of dexterous hands, the primary challenge is to synthesize stable and plausible dexterous grasping solutions~\cite{Rajeswaran2017LearningCD, Hasson2019Obman, Karunratanakul2020GraspingField, She2022LearningHR, Li2022GenDexGraspGD, Wang2022DexGraspNetAL, Turpin2022GraspD, Wei2022DVGGDV, Geng2023UniDexGrasp++ID, Xu2023UniDexGraspUR}. Some works~\cite{Hasson2019Obman, Liu2021SynthesizingDA, Wang2022DexGraspNetAL, Li2022GenDexGraspGD} employ optimization algorithms and utilize grasping metrics as objective functions, such as force closure~\cite{Dai2018SynthsisOp} for stability and penetration computation ~\cite{Dai2018SynthsisOp, Wang2022DexGraspNetAL} for collision detection. For example, Obman~\cite{Hasson2019Obman} utilizes GraspIt! ~\cite{Miller2004Graspit!} to produce feasible and collision-free grasps, while DexGraspNet~\cite{Wang2022DexGraspNetAL} leverages an differentiable force closure estimator to generate stable grasp results. These approaches are able to produce a large variety of physically feasible grasps, but without considering the reasonable and natural arrangement of results, thus not convenient for the subsequent manipulation tasks in real scenarios.

On the other hand, data-driven approaches~\cite{Hampali2019HOnnotateAM, Duan2021RoboticsDG, Liu2019GeneratingGP, Chao2021DexYCBAB, Wei2022DVGGDV, Li2022EfficientGraspAU, xu2024dexterous} have become popular, utilizing large annotated datasets to learn to generate plausible and human-like grasps. Many works~\cite{Taheri2020GRABAD, Karunratanakul2020GraspingField, Wei2022DVGGDV, Li2024SemGraspSG} train Variational Autoencoder networks (VAEs) to learn the grasp distribution and sample diversity grasp solution based on the given condition. Some others~\cite{Lu2023UGGUG, Weng2024DexDiffuserGD, Chang2024Text2GraspGS} utilize the recent diffusion model to sample dexterous grasps via the iterative denoising process. While data-driven approaches are capable of efficiently generating diverse and reasonable grasps, generating high-quality grasps remains a tough task.

To further enhance the grasping quality of the generated dexterous hands, some related works resort to intermediate representations to guide the grasping generation. Contact map~\cite{Brahmbhatt2019ContactDBAA} is a widely used representation with the vertex-wise contact information of the dexterous hands. Many approaches~\cite{Yang2020CPFLA, Brahmbhatt2019ContactGraspFM, Grady2021ContactOptOC, Jiang2021HandObjectCContactTTA, Li2022Contact2Grasp3G, Tse2022S2ContactGN, Liu2023ContactGenGC} generate reasonable hands based on the contact map prior information. In addition, FunctionalGrasp~\cite{Zhang2023FunctionalGraspLF} uses touch codes~\cite{Zhu2021TowardHG}, which record the contact status of each finger part w.r.t. the object's surface. These fine-grained intermediate representations are used for a following conditional generation or optimization-based grasping refinement. 
\begin{figure*}[!t]
   \includegraphics[width=1.0\linewidth]{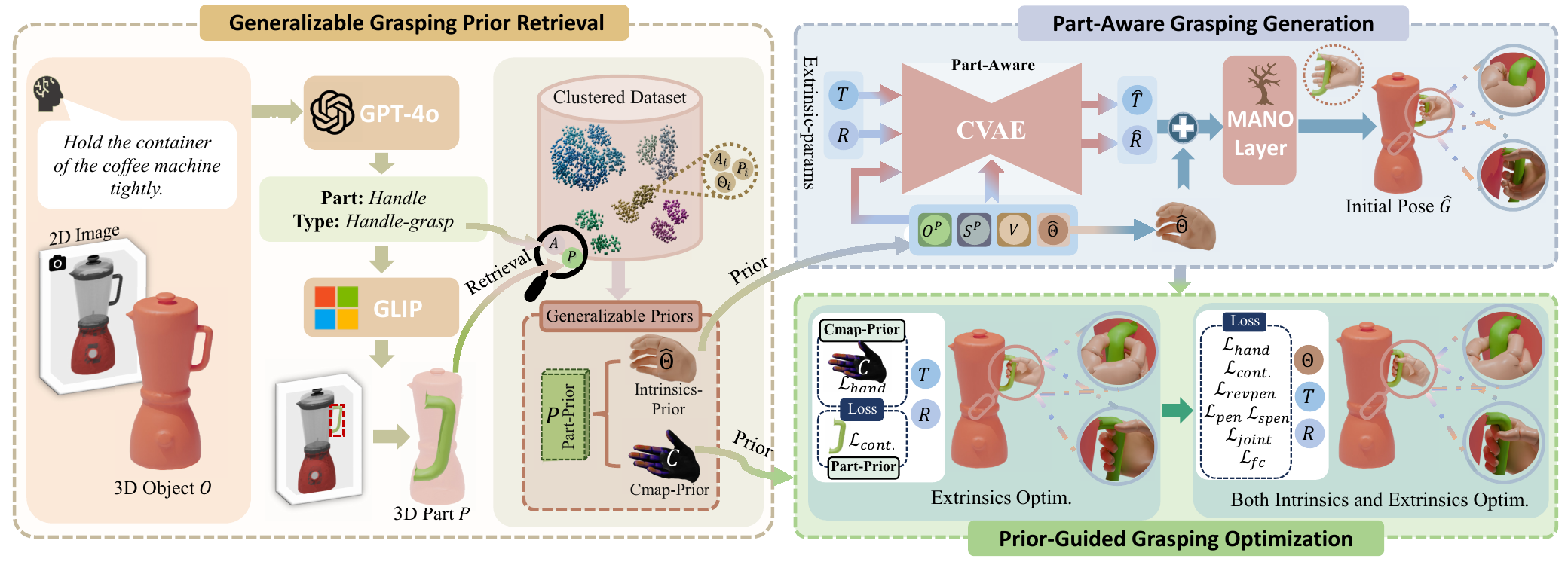}
   \caption{\textbf{The three-stage pipeline of G-DexGrasp.} (1) Generalizable grasping prior retrieval leverages pre-trained models to infer the fine-grained grasp configuration and retrieve relevant instances to form the grasping prior. (2) Part-aware grasping generation takes the target contact part and the retrieved prior as conditions to roughly initialize the hand. (3) Prior-guided grasping optimization takes the retrieved prior and other constraints as objective loss. It first optimizes the extrinsics only and then both extrinsic and intrinsic parameters.}
   \label{fig:pipeline}
   %\vspace{-13pt}
\end{figure*} 

%-------------------------------------------------------------------
\subsection{Task-Oriented Dexterous Grasping}

% One of the primary advantages of using dexterous hands is the high flexibility to conduct a large variety of manipulation tasks. In addition to the stability and plausibility purpose, more and more research studies task-oriented dexterous grasping, i.e. synthesizing the dexterous grasping that is friendly to the subsequent manipulation tasks.
One of the main advantages of dexterous hands is their high flexibility for performing diverse manipulation tasks. Beyond ensuring stability and plausibility, increasing attention has been given to task-oriented dexterous grasping—i.e., generating grasps that facilitate subsequent manipulation.

Some related works~\cite{Brahmbhatt2020ContactPoseAD, Taheri2020GRABAD, Yang2022OakInkAL} classify the task intents into several categories, such as ``use", ``pass", etc. They collect large amounts of grasping instances, including the task intent, target object, and dexterous hand pose, and train conditional generative networks to learn the distribution of task-oriented grasps. Moreover, AffordPose~\cite{Jian2023AffordPoseAL} collects the grasping instances and the fine-grained annotations, i.e. the part-level affordance labels such as ``twist" and ``handle-grasp". The affordance performs as a fine-grained classification of the task intent, thus producing more reasonable grasps for different objects and task intents.

The rapid development of natural language processing and the pre-trained large language models have enabled understanding and reasoning about task intent from natural language. Task2Grasp~\cite{Chang2024Text2GraspGS} uses the CLIP text encoder~\cite{Radford2021Clip} to process the task instruction and trains a diffusion model conditioned on the text encoding and the object point cloud to generate hand poses. By contrast, GraspAsYouSay~\cite{Wei2024DexGYGrasp} utilizes more detailed, finger-level instructions to guide grasp generation. SemGrasp~\cite{Li2024SemGraspSG} fine-tunes the pre-trained Multimodal Large Language Model (MLLM), i.e. LLaVA~\cite{Liu2023LLaVA}, to predict grasp tokens based on the task instruction and object point cloud, which are then decoded into hand parameters via a pre-trained hierarchical VAE. The concurrent work AffordDexGrasp~\cite{wei2025afforddexgrasp} generates generalizable affordances to assist grasping.

In light of the generalization of task-oriented dexterous grasping, the key is to infer the grasping arrangement for unseen objects w.r.t. the specified task instruction. Considering the large variation of object shapes and delicate grasping parameters, the performance of the above approaches is limited when directly transferred to deal with unseen objects. Inspired by the concept of generalizable parts~\cite{Tang2023GraspGPTLS, Geng2022GAPartNetCD, Tekden2023GraspTransferB} and the high-quality dexterous hand synthesis with part-level affordances~\cite{Jian2023AffordPoseAL}, we utilize the pre-trained large models to retrieve the fine-grained generalizable grasping prior and synthesize the reliable and semantically aligned dexterous hands for unseen objects.
\vspace{-5pt}

\section{Method}
\label{sec:method}

%=========================================================
\subsection{Overview}

Our problem follows the task-oriented dexterous grasping synthesis, where the input contains a task instruction $L$ in the form of natural language and a target object represented as 3D point cloud, i.e. $O \in \mathbb{R}^{N\times3}$. The output is the pose parameters of the synthesized dexterous hand, denoted as $G=(T,R,\Theta)$. $T \in \mathbb{R}^3$ and $R \in SO(3)$ are the extrinsic parameters representing the transformation and rotation of the global hand, while $\Theta \in \mathbb{R}^d$ is the intrinsic parameters representing angle values for the joints of the dexterous hand. We have $d=16$ with the articulated kinematic MANO hand model~\cite{Hasson2019Obman, Romero2017MANO}. The synthesized hand should be able to stably and anthropomorphically hold the target object and be semantically aligned with the specified task instruction. In this paper, we especially focus on \emph{generalizable dexterous grasping synthesis}. That is, the input task instructions and target objects are not limited within the distribution of the training data, but rather generalized to the diverse task instructions and unseen object categories.

We leverage the retrieval-augmented generation (RAG) strategy in our approach. To support the retrieval of relevant grasping instances, we pre-process the existing dataset as $\Omega=\{(L_{i},O_{i},G_{i},A_{i},P_{i},M_{i})\}_1^N$. Each data $\Omega_i$ includes the task instruction $L_{i}$, object model $O_{i}$, hand pose $G_{i}$, and the annotated affordance type $A_{i}$, contact part $P_{i}$ of the grasping instances, part segmentation $M_{i}$ of 3D objects.% The generalizable parts are of vital importance in the grasping prior retrieval and transfer to the unseen objects.

% Our approach operates in three stages. Specifically, as illustrated in Figure~\ref{fig:pipeline}, by analyzing the task instruction and target object, we first retrieve the relevant grasping instances and organize the generalizable grasping prior. The grasping prior represents the grasping knowledge for the specific scenario and is composed of the representative hand intrinsics and the contact map distribution, as well as the contact part of the target object. Next, we train a generative part-aware grasping network to estimate a reasonable extrinsic hand parameter conditioned on the specified intrinsics and contact part. Finally, taking the retrieved intrinsics and the generated extrinsics as the initial hand pose, we perform an iterative optimization to refine the dexterous hand with physical constraints and the retrieved grasping prior.
Our approach consists of three stages. As shown in Figure~\ref{fig:pipeline}, we first analyze the task instruction and target object to retrieve relevant grasping instances and organize a generalizable grasping prior, which includes representative hand intrinsics, contact map distribution, and the target’s contact part. Next, we train a part-aware generative network to estimate extrinsic hand parameters conditioned on the given intrinsics and contact part. Finally, using the retrieved intrinsics and generated extrinsics as the initial pose, we refine the hand via iterative optimization with physical constraints and the retrieved prior.

%=========================================================
\subsection{Generalizable Grasping Prior Retrieval}

%Given a task instruction and a 3D object, the first stage is to retrieve the relevant and generalizable grasping prior from the pre-processed dataset to guide the subsequent hand pose synthesis. 
In this stage, given the task instruction and the target object, we utilize the pre-trained Multimodal Large Language Model (MLLM) to infer the task-oriented and fine-grained grasping arrangement, i.e., the affordance type and contact part. Then we retrieve relevant grasping instances from the dataset and organize their distributions as the grasping prior. The prior is generalizable because both the affordance and the contact part are generalizable across object categories. The former indicates the rough hand intrinsics and the latter refers to the grasping region on the object surface.

\noindent\textbf{Task Instruction Analysis.} We first prompt the MLLM to determine the fine-grained grasping arrangement. It takes the task instruction $L$ and the rendered image $I$ of object $O$, and outputs the affordance type $A$ and the name of contact part $P$. The image is rendered from the selected viewpoint to show its shape and appearance as much as possible. In our prompt, we provide a set of pre-defined affordance types as well as their descriptions and restrict the MLLM to select one from them. For example, given the image of a \textit{kettle} and the instruction \textit{``could you please hold the kettle to pour some tea?"}, the MLLM should infer the affordance type \textit{``handle-grasp"} and the contact part \textit{``handle"}.

\noindent\textbf{Contact Part Localization.} Given the inferred part name, we need to localize the corresponding part region of the 3D object. We employ the pre-trained GLIP model~\cite{li2022grounded}, which takes the rendered image and the contact part name as input and outputs the detected 2D bounding boxes of the contact part. Then the pixels within the bounding boxes are projected to the 3D point cloud of the object, with the help of an additional depth map rendered from the same viewpoint. Due to the occlusion with the rendered image, we obtain some partial 3D parts from the detected bounding boxes. In our implementation, we utilize the segmented objects in the pre-processed dataset and vote from these partial parts to find the complete contact part $P$. For objects without pre-segmentation, an alternative solution is to fuse multi-view segmentation using existing methods~\cite{liu2023partslip, zhou2023partslip++}.

\noindent\textbf{Grasping Prior Retrieval.} To retrieve the relevant grasping instances from the dataset, we define the dissimilarity between grasping instances $\Omega_i$ and $\Omega_j$:
\begin{equation}
Dist(\Omega_{i},\Omega_{j})=D_A(A_i,A_j)+D_P(P_i,P_j)+D_{\Theta}(\Theta_i,\Theta_j),
\end{equation}
where $D_A$, $D_P$, and $D_{\Theta}$ measure the dissimilarity between the affordance types, contact parts, and the hand intrinsics, respectively. With $OBB$ representing the sorted size of the contact part's oriented bounding box, we have 
\begin{equation}
\centering
\begin{aligned}
D_A(A_i,&A_j)=\begin{cases}
\infty, A_i\neq A_j
\\
0, A_i=A_j
\end{cases}, 
\quad
D_{\Theta}=\left \| \Theta_i-\Theta_j \right \| ,
\\
&D_P(P_i,P_j)=||OBB(P_i)-OBB(P_j)||.
\end{aligned}
\label{Eq: OBB-based similarity}
\end{equation}
Based on this similarity, we perform K-means clustering on the dataset. Each clustering contains the grasping instances that belong to the same affordance type and have similar contact part shapes and intrinsic hand poses.  %~\Jan{It is worth noting that our similarity metric is easily replaceable. For example, we try BPS~\cite{Prokudin2019EfficientBPSLO} to encode 3D parts as features and measure with cosine similarity, the experimental results (in the \emph{suppl. material.}) are close to OBB-based methods. So, we choose the feasible method for implementation.}

Instead of the other retrieval-augmented generation approaches that only utilize the most similar instance, we organize the grasping knowledge of each cluster as a generalizable grasping prior $\Gamma=(P, \hat{\Theta}, C)$ that includes the localized contact part $P$, the representative intrinsic pose $\hat{\Theta}$, and the contact map distribution $C$ for hand. The representative pose $\hat{\Theta}$ is the nearest neighbor instance to the mean of the clustering. The hand contact map distribution $C$ refers to the mean and standard deviation of the Gaussian distribution of the contact maps in the cluster, where a hand contact map reflects the nearest distance of each mesh vertex of the hand to the object surface~\cite{Grady2021ContactOptOC}. Note that the computation of generalizable grasping prior is scalable, meaning that we can obtain convincing priors from a small-scale dataset and can flexibly extend it with more types of grasp instances.

%=========================================================
\subsection{Part-Aware Grasping Generation Network}

The second stage aims to estimate the rough hand parameters $\hat{G}=(\hat{T},\hat{R},\hat{\Theta})$ based on the given object and the inferred grasping arrangement. We directly take the retrieved representative hand pose $\hat{\Theta}$ as the intrinsic parameters and train a network to predict corresponding extrinsic parameters, i.e. $\hat{T}$ and $\hat{R}$, in the coordinate system origined at contact part center. In order to generalize to unseen objects, we design the part-aware grasping generation network, supervised by the ground-truth hand extrinsics from the training set without intrinsic parameters.

\begin{figure}[!t]
   \includegraphics[width=1.01\linewidth]{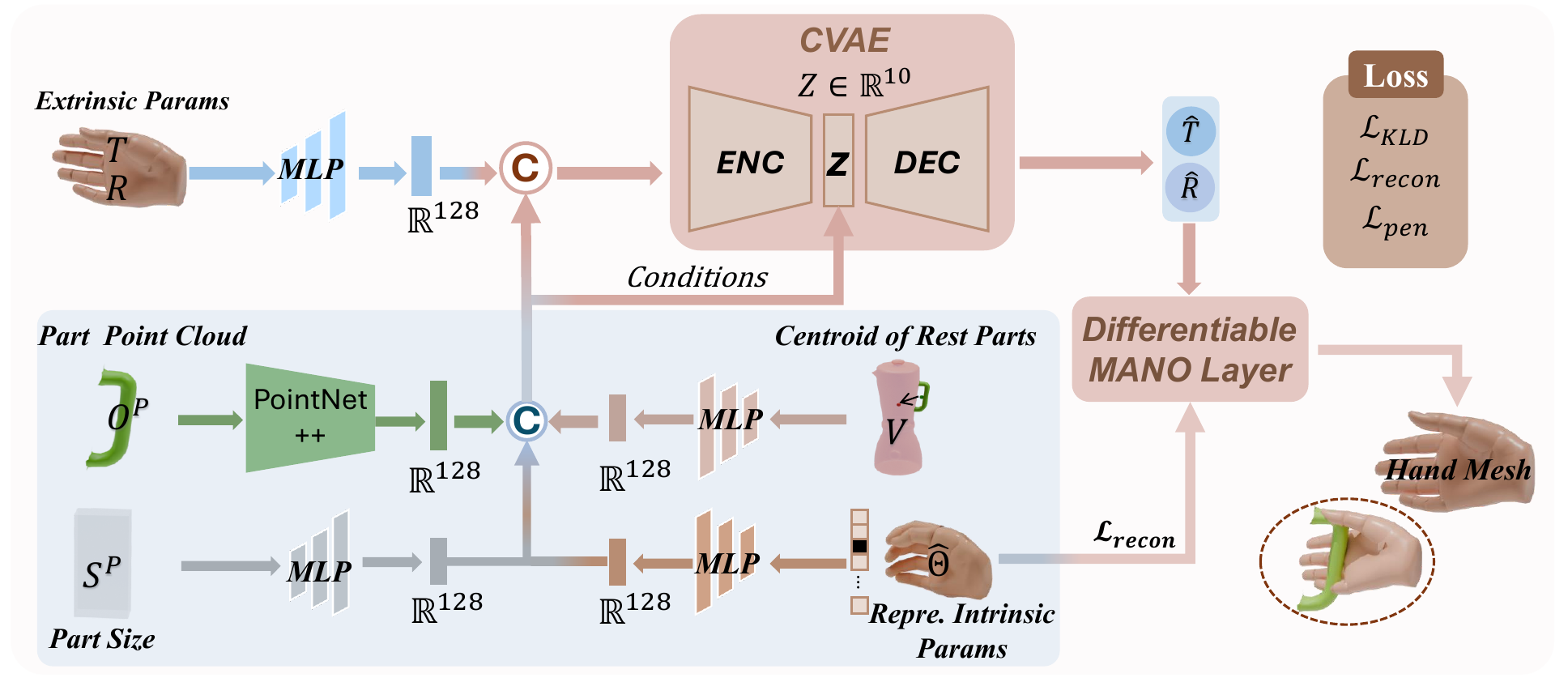}
   \caption{Network architecture of part-aware generation network.}
   \label{fig:cvae-network}
\end{figure}

The network is implemented as a conditional variational auto-encoder, as shown in Figure~\ref{fig:cvae-network}. The input condition includes contact part point cloud $O^{P}$, part size $S^{P}$, representative intrinsic parameters $\hat{\Theta}$, centroid $V$ of the rest of the object. The point cloud $O^{P}$ is normalized into a unit sphere with a scaling factor, while the part size $S^{P}$ is the axis-aligned bounding box size% of the original part point cloud
. The centroid $V$ is defined as the center location of the object excluding the contact part, w.r.t. the center of the contact part. It prevents a large overlap between the predicted hand and the object.

To encode the input conditions, for the part point cloud, we concatenate each point with its normal vector and the scaling factor, and use PointNet++~\cite{qi2017pointnet++} to produce a $128$-dimensional latent vector. The representative intrinsic parameters $\hat{\Theta}$ are represented as the one-hot vector of its corresponding cluster in the dataset. This one-hot vector, centroid $V$ and part size $S^{P}$ are encoded as $128$-dimensional latent vectors using linear layers, respectively. The latent vectors are concatenated to compose the latent condition vector. The CVAE network takes the condition vector and learns to reconstruct extrinsic parameters, i.e. $T$ and $R$, which are fed into a differentiable MANO layer~\cite{Hasson2019Obman} to produce the mesh of the corresponding dexterous hand. 

The loss function of the grasping generation network is  
\begin{equation}
\mathcal{L}= \mathcal{\lambda}_{KLD}\mathcal{L}_{KLD} + \mathcal{L}_{recon} + \mathcal{\lambda}_{pen}\mathcal{L}_{pen},
\end{equation}
where $\mathcal{L}_{KLD}$ is the KL-divergence loss of the CVAE network to align the learned distribution with a Gaussian distribution. The reconstruction loss $\mathcal{L}_{recon}$ is the mean square loss between the reconstructed hand and the ground-truth
\begin{equation}
\mathcal{L}_{recon} = \mathcal{\lambda}_{V}·||{\hat{\mathcal{V}}}-\mathcal{V}||+\mathcal{\lambda}_{R}·||\hat{R}-R|| + \mathcal{\lambda}_{T}·||\hat{T}-T||,
\end{equation}
where $\hat{\mathcal{V}}$, $\hat{R}$, $\hat{T}$ represent the predicted mesh vertices, global rotation, and global translation of the hand, while $\mathcal{V}$, $R$, $T$ are the ground-truth. The penetration loss $\mathcal{L}_{pen}$ penalizes the overlapping between the dexterous hand and the object
\begin{equation}
\mathcal{L}_{pen}=\sum_i\max(0,(O_{i}-\hat{\mathcal{V}}_i) \cdot \mathbf{n}^O_{i}-c_{pen}),
\label{eq_pen}
\end{equation}
where $O_i$ and $\mathbf{n}^O_{i}$ denote the object mesh vertex and its corresponding normal vector, $\hat{\mathcal{V}}_i$ is the nearest hand vertex to $O_i$, $c_{pen}$ is a pre-determined threshold.

%=========================================================
\subsection{Prior-Guided Grasping Optimization}
It is difficult to have a network to directly generate precise grasping parameters, especially for unseen object categories. Therefore, we use the retrieved generalizable grasping prior as the guidance of a differentiable grasp optimizer to refine the roughly estimated hand parameters.

Given the initial dexterous hand parameters $\hat{G}=(\hat{T},\hat{R},\hat{\Theta})$, we use the differentiable MANO layer to obtain the hand mesh and compute the objective function. The objective function consists of a variety of terms from three aspects. One is the \emph{retrieved grasping prior guidance}, which is composed of $\mathcal{L}_{contact}$ to encourage the hand to contact with the target part and $\mathcal{L}_{cmap}$ to have the contact map conforming to the retrieved distribution. We have

\begin{equation}
\begin{aligned}
&\mathcal{L}_{contact}=\frac{1}{n}\sum_{j=1}^{n}C_{j}\underset{i \in N}{\min} \|\ \hat{\mathcal{V}}_j-O^{P}_{i} \|, 
\\
&\mathcal{L}_{cmap}=\frac{1}{n}\sum_{j=1}^{n}e^{-\sigma_j}\|\hat{C}-C\|,
\end{aligned}
\end{equation}
where $C$ is the averaged hand contact map of the retrieved grasp cluster and $C_j$ is the averaged contact map value of the $j$th hand vertex. $\sigma_j$ is the standard deviation of the contact maps of the retrieved grasp cluster at the $j$-th vertex. $\hat{C}$ is the contact map of the predicted hand. $\hat{\mathcal{V}}_j$ and $O^{P}_{i}$ are the vertices of the hand and contact part. Following the prior works, the second is the \emph{physical feasibility constraints} of the grasping, including the differentiable force closure term $\mathcal{L}_{fc}$~\cite{Wang2022DexGraspNetAL} to enhance the grasping stability, the penetration term $\mathcal{L}_{pen}$~\cite{Grady2021ContactOptOC} (Eq.~\ref{eq_pen}) and $\mathcal{L}_{revpen}$ to avoid hand-object overlapping, i.e.
\begin{equation}
\mathcal{L}_{revpen}=\sum_i\max(0,(\hat{\mathcal{V}_i}-O_{i}) \cdot \mathbf{n}^{\hat{\mathcal{V}}}_{i}-c_{revpen}),
\label{eq_pen}
\end{equation}
where $\hat{\mathcal{V}_i}$ and $\mathbf{n}^{\hat{\mathcal{V}}}_{i}$ denote the predicted hand mesh vertex and its normal vector, $O_{i}$ is the nearest object vertex to $\hat{\mathcal{V}_i}$, $c_{revpen}$ is a pre-determined threshold.
The third is the \emph{human-like grasping constraints}, containing $\mathcal{L}_{joint}$~\cite{Wang2022DexGraspNetAL} to restrict each angle within the range of human joint and $\mathcal{L}_{spen}$~\cite{Zhu2021TowardHG} to avoid self-penetration of the hand.

Our optimization contains two sub-processes. Specifically, the first aims to refine the extrinsic parameters while preserving the intrinsic hand pose. It only utilizes the retrieved grasping prior guidance as the objective function. The goal is to quickly move the hand to its target position and orientation, rather than struggling at an incorrect local minimum. The second stage fine-tunes all the parameters to obtain the final dexterous hand $G=(T,R,\Theta)$, where all the above terms are combined with their weights as the objective function. We use Adam optimizer for the two processes, where the first uses a learning rate of 0.005 and the second 0.001. Each stage contains 700 iterations of updates.

\section{Experiments}
\label{sec:exp}

\begin{table*}
\centering
% \begin{tabular}{l|ccc|cc}
\resizebox{0.97\textwidth}{!}{
\begin{tabular}{l|p{3cm}<{\centering}p{2.5cm}<{\centering}p{1.8cm}<{\centering}|p{1.9cm}<{\centering}p{2.3cm}<{\centering}}
\hline
\multirow{2}{*}{Experiments} & \multicolumn{3}{c|}{Quality and Stability}                                                                         & \multicolumn{2}{c}{Semantic Alignment}                                       \\ 
\cline{2-6}
                             & Solid.Intsec.Vol($cm^3$)$\downarrow$ & Penet.Depth($cm$)$\downarrow$        & Sim.Dis($cm$)$\downarrow$            & Part Acc.$(\%)$                     & Percep.Score$(\%)$                    \\ 
\hline
AffordPose~\cite{Jian2023AffordPoseAL}       & 6.86/10.55                              & 0.96/1.21                            & 10.66/9.93                          & 81.0/42.5                            & 3.84/1.72  \\
GrabNet$^\dag$~\cite{Taheri2020GRABAD}       & 8.76/11.07                              & 1.08/1.47                           & 10.58/9.62                          & 75.6/45.5                            & 3.67/1.63  \\
GraspTTA$^\dag$~\cite{Jiang2021HandObjectCContactTTA} & 5.34/7.98                    & 0.89/0.92                           & 8.17/7.61                           & 74.3/46.5                            & 2.03/0.88  \\
Ours                                        & \cellsecond 3.13/2.94                    & \cellsecond 0.82/0.29                & \cellsecond5.39/4.27               & \cellsecond 82.6/71.6                 & \cellsecond 3.72/3.93  \\
Ours$^*$                                    &\cellfirst 3.18/2.04                   &\cellfirst 0.87/0.22               &\cellfirst 4.93/4.34                & \cellfirst 89.74/96.3                 & \cellfirst 3.94/4.44 \\
\hline
\end{tabular}
}
\caption{Quantitative comparison on the seen$/$unseen object categories. GrabNet$^\dag$ and GraspTTA$^\dag$ use BERT to encode task instructions. Ours uses pre-trained models to infer and localize contact parts. Ours$^*$ uses ground-truth part segmentation with inferred contact part selection. Best and second-best results are marked in \textfirst{\textbf{red}} and \secondtext{\textbf{green}}, averaged over seen and unseen cases.}
\label{tab:comparison_results}
\end{table*}

%=========================================================
\subsection{Experiment Settings}

\noindent \textbf{Dataset.} We use the AffordPose dataset~\cite{Jian2023AffordPoseAL}, which provides dexterous hand data, 3D object shapes, and part-level affordance labels for each grasp. It includes 8 interaction types across 13 object categories. We select five grasp-related interactions—\emph{handle-grip}, \emph{twist}, \emph{wrap-around-grip}, \emph{base-support}, and \emph{trigger-squeeze}—and split the data 8:1:1 for training, validation, and testing. The training set is used for retrieval and generation network training, while the test set validates performance on \emph{seen} categories.

We select novel objects to test generalization, including 92 objects from 11 categories in OakInk~\cite{Yang2022OakInkAL} (e.g., frying pan, bowl), 81 objects from 13 categories in PartNet-Mobility~\cite{Xiang2020SAPIEN} (e.g., suitcase, coffee machine), and 126 objects from 14 categories in 3D-Net~\cite{Wohlkinger20123DNetLO} (e.g., broom, umbrella), forming our \emph{unseen} test set.

We preprocess datasets to prepare task instructions and part segmentations. A pre-trained MLLM generates task instructions for each affordance type and contact part. Specifically, we render the object image highlighting the contact part in green and prompt the MLLM: \emph{``your task is to create specific, indirect instructions or commands, commanding the robot to perform $\{$affordance type$\}$ by interacting with the green part after carefully reasoning the command context".} For part segmentation, we take the provided ground-truth from the datasets and manually annotate on the point clouds of other objects without the provided ground-truth.

\noindent \textbf{Metrics.}
We adopt the commonly-used metrics~\cite{Jiang2021HandObjectCContactTTA, Jian2023AffordPoseAL} to assess the plausibility and the alignment with the task instruction of the generated grasps. For plausibility, we measure \textbf{penetration depth} (Penet.Depth, $cm$) and \textbf{solid intersection volume} (Solid.Intsec.Vol, $cm^3$) of the overlap between the object and hand meshes. %The depth metric reflects the \color{red}maximum or mean distance \color{black}from hand mesh vertices to the object surface when penetration occurs, while simulation displacement provides further insight into interaction realism. 
The \textbf{simulation displacement} (Sim.Dis. $cm$) measures the displacement of object's center of mass over a period of time with the generated grasps, indicating the grasp stability in the simulation environment~\cite{Jiang2021HandObjectCContactTTA, Hasson2019Obman}. %In simulation, forces exerted on the fingertips, which are correlated with the fingertip penetration volume, act to counterbalance the weight of the object. The grasp stability is indicated by the displacement of the object's center of mass over a set period, during which the hand's configuration and position remain constant. We calculate the mean of the simulation displacement for test data, excluding some large objects such as \emph{Storage Furniture}, with lower displacements indicating greater stability. 
\noindent For semantic alignment, we follow the protocol of~\cite{Jian2023AffordPoseAL}  for the \textbf{part accuracy} (Part Acc. $\%$) metric to verify whether the generated hand contact with the object at the ground-truth contact part. We also assess \textbf{Perceptual Score} (Percep.Score $\%$) for which we conduct a user study of 20 groups of results with 30 participants to judge the quality and task alignment of generated grasps. 

%if the predicted hand’s contact region aligns with the ground-truth part conditions, following~\cite{Jian2023AffordPoseAL}. The contacting region is defined as the area with the highest concentration of contact points ($ dis_{o2h} \leq \alpha = 0.004m $); if there's no contact point under this threshold, we gradually increase $ \alpha $ by $ step=0.001m $ until the contact points appear or $ \alpha $ reaches $0.01$. If there's no contact point when $\alpha = 0.01$, we directly mark the result as a wrongly predicted hand. We randomly selected 10 objects and assessed \textbf{Perceptual Score}(Percep.Score $\%$) using a user study and a multi-modal large language model (MLLM), focusing on hand naturalness, human-likeness, and task alignment.

\noindent \textbf{Implementation.}
We invoke GPT-4o as the pre-trained MLLM to infer the affordance type and contact part of the dexterous grasping. We sample $N=2048$ points from the object surface as the input object points. During training, we use the Adam optimizer with a learning rate of 1e-3 to train the network for 400 epochs. 
% We set the batch size as 90. 
The loss weights are  $\mathcal{\lambda}_{V}=100$, $\mathcal{\lambda}_{R}=5$, $\mathcal{\lambda}_{T}=30$, $\mathcal{\lambda}_{KLD}=10$, $\mathcal{\lambda}_{pen}=1$.

\begin{figure*}[!t]
   \includegraphics[width=1.0\linewidth]{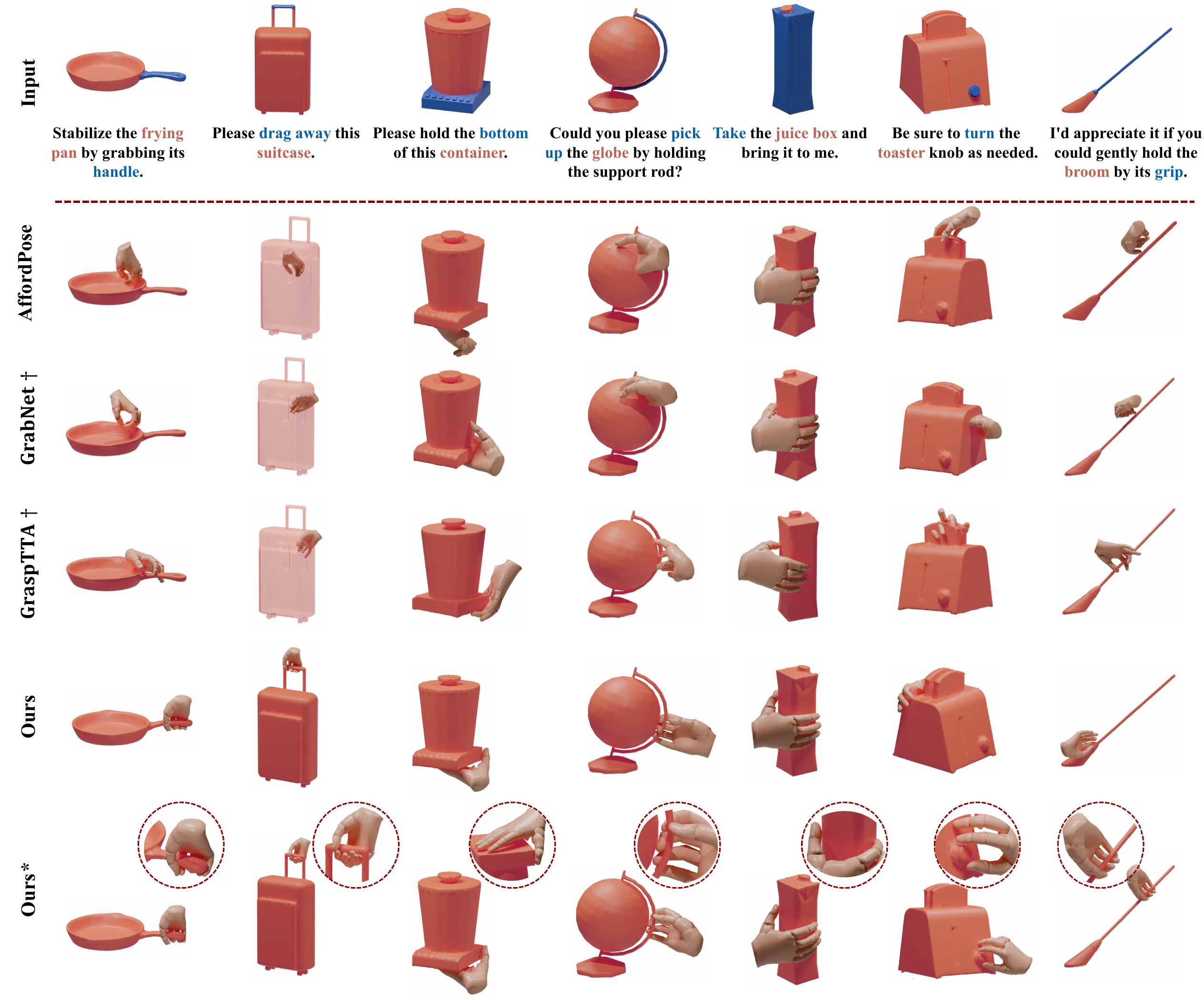}
   \caption{Qualitative comparison. Given object from unseen categories and language-based task instruction, our approach exhibits remarkable performance in the generalization. Note that the contact parts are highlighted in the first row for visualization, not as input.}
   \label{fig:comparison results}
   \vspace{-5pt}
\end{figure*} 
%=========================================================
\subsection{Comparisons}

We compare against the state-of-the-art methods, which focus on taking the affordance type labels or the embedding of textual task descriptions as conditions for dexterous grasping synthesis. Note that AffordPose~\cite{Jian2023AffordPoseAL} and OakInk~\cite{Yang2022OakInkAL} actually use the same GrabNet network~\cite{Taheri2020GRABAD} but training data with different sets of task labels. So we select AffordPose as their representative. In addition, with the absence of released code of Text2Grasp~\cite{Chang2024Text2GraspGS}, we follow their method which uses pre-trained BERT~\cite{kenton2019bert} to encode the task descriptions and select the existing network architectures GrabNet~\cite{Taheri2020GRABAD} and GraspTTA~\cite{Jiang2021HandObjectCContactTTA} to examine the performance on task-oriented dexterous grasping, denoted as GrabNet$^\dag$ and GrabTTA$^\dag$. We also provide two different versions of our approaches. One is our full method which uses the pre-trained model to localize contact parts, while the other, denoted as Ours*, uses the ground-truth part segmentation with inferred contact part selection.

%Existing works for task-oriented dexterous grasping task focus on using fine-grained annotations, such as AffordPose and OakInk, or detailed task instructions in the form of natural language descriptions, such as Text2Grasp, to guide the network to learn the hand pose distributions. Among them, AffordPose and OakInk use the network architecture of GrabNet, while the more recent works have not released the code yet. Therefore, we adopt the network architecture of GrabNet and GraspTTA, where the former is widely used in many related works while the latter is a recent approach for task-agnostic dexterous grasping, and use different task intent conditions, including the fine-grained affordance labels and task instruction descriptions, to train the networks to learn the task-oriented grasping. In short, we compare against the competitive methods denoted as GrabNet (Inst.), GrabNet (ALabel), GraspTTA (Inst.), GraspTTA (ALabel). Inst. denotes the task instruction condition embedded via the pre-trained BERT. ALabel stands for the affordance label condition embedded as latent vectors via linear layers.

The qualitative and quantitative evaluations are reported in Table~\ref{tab:comparison_results} and Figure~\ref{fig:comparison results}. The first two rows of Table~\ref{tab:comparison_results} use the same network architecture but different conditions, i.e. object point cloud and either the affordance type (first row) or the task description (second row). It shows that the task descriptions act better than affordance types in guiding the dexterous hand to contact the correct region on the unseen objects in our test set (see the semantic alignment metrics). Due to the same network architecture, they exhibit similar performance in terms of grasping quality and stability. By contrast, GraspTTA$^\dag$ shows better stability but similar semantic alignment with GraspNet$^\dag$, due to the same task description condition but a more advanced network.

However, more importantly, our approach achieves a remarkable improvement in the performance on unseen object categories, thanks to our retrieval-augmented grasping generation strategy. In addition, when utilizing the ground-truth part segmentations, i.e. ours*, all the metrics exhibit a further improvement, especially the semantic alignment. It adequately validates the transferability of our generalizable grasping prior on unseen object categories and the potential of our approach with the further development of open-vocabulary part segmentation approaches.

%=========================================================

\subsection{Ablation Study}

\begin{figure*}[!t]
\centering
   \begin{overpic}[width=1.0\linewidth,tics=10]{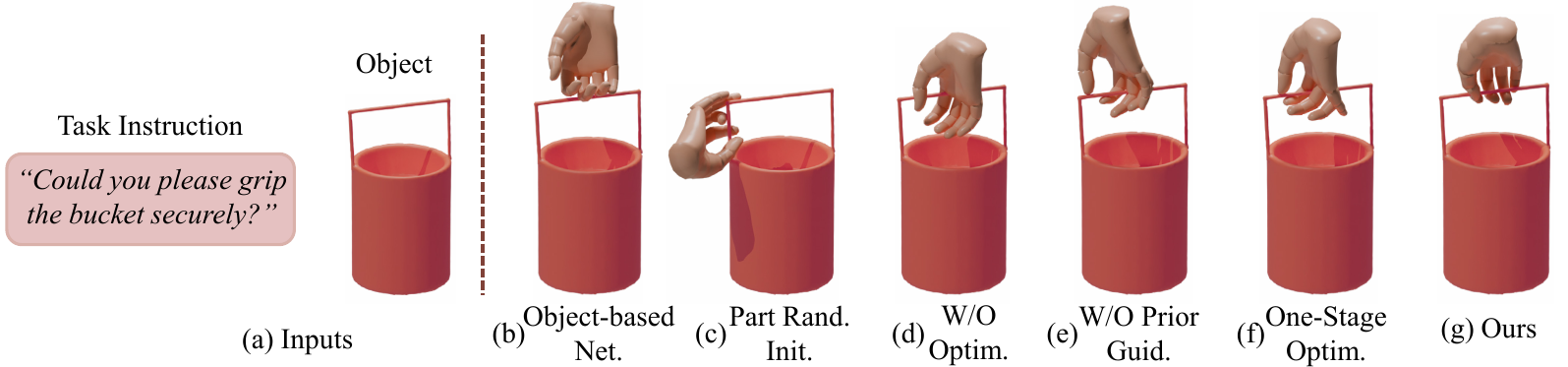}
% \put(17,-1){\small (a) Inputs}
% \put(33,-1){\small (b) \parbox[c]{2cm}{\centering Object-based \\ Net.}}
% \put(47,-1){\small (c) \parbox[c]{2cm}{\centering Part-aware \\ Rand. Init.}}
% \put(62,-1){\small (d) \parbox[c]{4cm}{W/O \\Optim.}}
% \put(70,-1){\small (e) \parbox[c]{4cm}{W/O Prior\\ Guid.}}
% \put(82,-1){\small (f) \parbox[c]{4cm}{One-Stage\\ Optim.}}
% \put(92,-1){\small (g) Ours}
\end{overpic}
\caption{Qualitative evaluation of ablation study. Compared to our results, (b) and (c) validate the importance of using a part-aware network for hand initialization, while (d), (e), and (f) reveal the effectiveness of the prior-guided optimization to produce stable grasps.}
   \label{fig:ablation results}
   \vspace{-5pt}
\end{figure*}

We validate our key designs, which are of vital importance to correctly and efficiently exploit the retrieved information for dexterous hand generation. Qualitative and quantitative evaluations are reported in Figure~\ref{fig:ablation results} and Table~\ref{table: ablation results}. 

\begin{table*}
\centering
\resizebox{0.94\textwidth}{!}{
\begin{tabular}{l|ccc|cc}
\hline
\multirow{2}{*}{Experiments} & \multicolumn{3}{c|}{Quality and Stability}                                                                         & \multicolumn{2}{c}{Semantic Alignment}                                       \\ 
\cline{2-6}
                             & Solid.Intsec.Vol($cm^3$)$\downarrow$ & Penet.Depth($cm$)$\downarrow$        & Sim.Dis($cm$)$\downarrow$            & Part Acc.$(\%)$                     & Percep.Score$(\%)$                    \\ 
\hline
Object-based Net.            & 10.06                                & 0.66                                & 8.64                                 & 51.17                                 & 1.43                                  \\
Part Rand. Init.             & \cellfirst 1.34                     & \cellfirst 0.21                      & 6.40                                 & 57.85                                 & 1.04                                  \\ 
\hline
W/O Optim.                   & 17.43                                & 1.55                                 & \cellfirst 3.3               & 66.88                                 & 1.92                                  \\
W/O Prior Guid.              & 9.13                                 & 0.82                                 & \cellsecond 3.65              & 68.22               & 2.30                \\
One-Stage Optim.             & 3.18                     & \cellsecond 0.28                     & 5.02                          & \cellsecond 69.23          & \cellsecond 2.51                                  \\ 
\hline
Ours                         & \cellsecond 2.94                     & 0.29               & 4.27                                 & \cellfirst 71.6         & \cellfirst 3.93  \\
\hline
\end{tabular}
}
\caption{Quantitative evaluation of our ablation study to validate the importance of the retrieved prior in the grasping synthesis pipeline.}
\label{table: ablation results}
\end{table*}

\textbf{How does the retrieved information assist the grasping generation?} We design two different variants of our part-aware grasping generation network, named as \emph{Object-based Net.} and \emph{Part Rand. Init}. The former uses an object-conditioned network and the latter uses a random initialization around the target part, to replace our part-aware network which generates initial extrinsic parameters for the subsequent optimization. Specifically, compared to our full approach, the object-conditioned network removes the part-related conditions, i.e. $V^*$ and $S^{P^*}$, and uses the object point cloud to replace part point cloud $O^{P^*}$. On the other hand, for \emph{Part Rand. Init.}, we follow the method of DexGraspNet~\cite{Wang2022DexGraspNetAL} to initialize the extrinsic parameters to place the dexterous hands around the contact part.
%initialize extrinsic parameters according to the method of dexgrapnet[].The displacement space of the hand is offset along the direction of the object normal vector on the expanded convex hull surface, and the offset range is $[0.015 m, 0.025 m]$. The rotation space is rotated in the direction of the palm facing the object, and the local rotation is performed around the direction of the normal vector perpendicular to the palm. The global rotation angle range is $[0,\frac{\pi}{2}]$, and the local rotation angle range is$[-\frac{\pi}{6},\frac{\pi}{6}]$
%on each object mesh, we first take its convex hull, then push every vertex of the hull away from the origin by 0.025m to obtain the inflated convex hull. Next, we sample a random point $p$ on the surface of the inflated convex hull, and compute the direction vector from $p$ to its nearest point on the original object mesh, then jitter this direction vector within a cone and get $\vec{n}$. Finally, the hand is moved to $p$, and rotated to face the same direction as $\vec{n}$, then push away from the object mesh along $\vec{n}$ by a random distance, and rotated around$\vec{n}$ randomly.In this way, we get the initialization internal parameter$T^*$,$R^*$.}

The first two rows of Table~\ref{table: ablation results} report the results. We can see that the object-based network performs much worse on all the metrics. Because the terrible extrinsic initializations often cause dexterous hands with heavy penetration with the object, making it hard for the following optimization to refine them to stable and reasonable grasps. By contrast, part-aware random initialization heuristically places the dexterous hands around the contact part, thus resulting in good performance in terms of quality and stability. However, the randomness affects the robustness of the approach, causing slightly worse simulation and part accuracy. In summary, these baselines validate the necessity of using a part-aware network to initialize extrinsic parameters.

%i.e. grasping prior retrieval, part-aware intrinsics generation, prior-guided grasping optimization. For each of them, we select the alternative option to conduct the baseline experiments. 

\textbf{How does the retrieved information improve the refinement grasping optimization?} We conduct the three alternative options w.r.t. our prior-guided grasping optimization. The experiment \emph{W/O Optim.} removes the optimization and directly takes the initialized dexterous hands as the final outputs. The experiment \emph{W/O Prior Guid.} uses the same retrieval-assisted dexterous hand initialization as our full approach, but doesn't use any retrieved information during the optimization process. More specifically, it removes $\mathcal{L}_{cmap}$ and replace the $\mathcal{L}_{contact}$ with the $\mathcal{L}_{close}$ term in~\cite{hang2024dexfuncgrasp}, 
which only encourages the fingers of dexterous hand to contact with the object without any information from the retrieved contact maps. The experiment \emph{One-Stage Optim.} exploits the retrieved prior as we did, but optimizes both the intrinsic and extrinsic parameters in one single stage. 

The third to the fifth row of Table~\ref{table: ablation results} show their results. Without optimization, although the hands often approach the correct parts, they exhibit heavy penetration with the object and cannot hold the object stably. Without the prior guidance in optimization, it often produces unnatural grasps with larger penetration and sometimes slide to contact with other parts. By contrast, the one-stage optimization can produce good results for many cases, but would fail when the initialized extrinsic parameters cause opposite rotations or large distances to the object. In summary, these baselines validate the necessity of leveraging the retrieved prior as the objective of the optimization to guarantee the quality and semantic alignment of the refined dexterous grasps.

\section{Conclusion}
We present G-DexGrasp which exploits the retrieval-augmented generation for dexterous grasping synthesis on unseen object categories and diverse task instructions. The key is to retrieve the generalizable grasping prior and take it as the fine-grained guidance for the subsequent generation and refinement optimization. Through extensive experiments, we validate the importance of leveraging a part-aware network to generate reasonable rough dexterous hands, and taking the retrieved prior as objective in a two-stage refinement optimization. Our results exhibit remarkable performance in terms of grasping quality and stability, as well as semantic alignment with task instructions.

However, our approach requires an open-vocabulary part segmentation for the unseen objects, which, though not the focus of this work, limits the performance of the synthesized dexterous hands. In addition, extending the pre-processed dataset with a wider range of task-oriented hand-object interaction instances, especially for those tasks demanding delicate dexterous hand postures, would further stimulate more manipulation tasks in various scenarios.  
 
%Better part segmentation, more manipulation tasks...

\section*{Acknowledgments.} This work was supported by National Natural Science Foundation of China (62272082, 62302269, 62027826, 62322207), the Excellent Young Scientists Fund Program (Overseas) of Shandong Province (2023HWYQ-034), Shenzhen University Natural Sciences 2035 Program (2022C007), and the joint Fund General Project of Liaoning Provincial Department of Science and Technology (2024-MSLH-352).

% Liu: 项目名称：面向机器人智能抓取任务的几何感知方法研究: National Natural Science Foundation of China (62272082)
% Li: National Natural Science Foundation of China (62302269)
% Liu: 面向恶劣海洋环境的具身智能系统构建与应用验证: National Natural Science Foundation of China (62027826), 
% Hu: National Natural Science Foundation of China (62322207)
% Li: Manyi Li is supported by the Excellent Young Scientists Fund Program (Overseas) of Shandong Province (grant no.2023HWYQ-034).
% Jian: the joint Fund General Project of Liaoning Provincial Department of Science and Technology (2024-MSLH-352)

{
    \small
    \bibliographystyle{ieeenat_fullname}
    \bibliography{main}

\begin{thebibliography}{4}
\providecommand{\natexlab}[1]{#1}
\providecommand{\url}[1]{\texttt{#1}}
\expandafter\ifx\csname urlstyle\endcsname\relax
  \providecommand{\doi}[1]{doi: #1}\else
  \providecommand{\doi}{doi: \begingroup \urlstyle{rm}\Url}\fi

\bibitem[Deitke et~al.(2023)Deitke, Schwenk, Salvador, Weihs, Michel, VanderBilt, Schmidt, Ehsani, Kembhavi, and Farhadi]{deitke2023objaverse}
Matt Deitke, Dustin Schwenk, Jordi Salvador, Luca Weihs, Oscar Michel, Eli VanderBilt, Ludwig Schmidt, Kiana Ehsani, Aniruddha Kembhavi, and Ali Farhadi.
\newblock Objaverse: A universe of annotated 3d objects.
\newblock In \emph{Proceedings of the IEEE/CVF conference on computer vision and pattern recognition}, pages 13142--13153, 2023.

\bibitem[Jiang et~al.(2021)Jiang, Liu, Wang, and Wang]{Jiang2021HandObjectCContactTTA}
Hanwen Jiang, Shaowei Liu, Jiashun Wang, and Xiaolong Wang.
\newblock Hand-object contact consistency reasoning for human grasps generation.
\newblock \emph{2021 IEEE/CVF International Conference on Computer Vision (ICCV)}, pages 11087--11096, 2021.

\bibitem[Liu et~al.(2025)Liu, Uy, Xiang, Su, Fidler, Sharp, and Gao]{liu2025partfield}
Minghua Liu, Mikaela~Angelina Uy, Donglai Xiang, Hao Su, Sanja Fidler, Nicholas Sharp, and Jun Gao.
\newblock Partfield: Learning 3d feature fields for part segmentation and beyond.
\newblock \emph{arXiv preprint arXiv:2504.11451}, 2025.

\bibitem[Prokudin et~al.(2019)Prokudin, Lassner, and Romero]{Prokudin2019EfficientBPSLO}
Sergey Prokudin, Christoph Lassner, and Javier Romero.
\newblock Efficient learning on point clouds with basis point sets.
\newblock \emph{2019 IEEE/CVF International Conference on Computer Vision (ICCV)}, pages 4331--4340, 2019.

\end{thebibliography}


\begin{thebibliography}{53}
\providecommand{\natexlab}[1]{#1}
\providecommand{\url}[1]{\texttt{#1}}
\expandafter\ifx\csname urlstyle\endcsname\relax
  \providecommand{\doi}[1]{doi: #1}\else
  \providecommand{\doi}{doi: \begingroup \urlstyle{rm}\Url}\fi

\bibitem[Brahmbhatt et~al.(2019{\natexlab{a}})Brahmbhatt, Ham, Kemp, and Hays]{Brahmbhatt2019ContactDBAA}
Samarth Brahmbhatt, Cusuh Ham, Charles~C. Kemp, and James Hays.
\newblock Contactdb: Analyzing and predicting grasp contact via thermal imaging.
\newblock \emph{2019 IEEE/CVF Conference on Computer Vision and Pattern Recognition (CVPR)}, pages 8701--8711, 2019{\natexlab{a}}.

\bibitem[Brahmbhatt et~al.(2019{\natexlab{b}})Brahmbhatt, Handa, Hays, and Fox]{Brahmbhatt2019ContactGraspFM}
Samarth Brahmbhatt, Ankur Handa, James Hays, and Dieter Fox.
\newblock Contactgrasp: Functional multi-finger grasp synthesis from contact.
\newblock \emph{2019 IEEE/RSJ International Conference on Intelligent Robots and Systems (IROS)}, pages 2386--2393, 2019{\natexlab{b}}.

\bibitem[Brahmbhatt et~al.(2020)Brahmbhatt, Tang, Twigg, Kemp, and Hays]{Brahmbhatt2020ContactPoseAD}
Samarth Brahmbhatt, Chengcheng Tang, Christopher~D. Twigg, Charles~C. Kemp, and James Hays.
\newblock Contactpose: A dataset of grasps with object contact and hand pose.
\newblock \emph{ArXiv}, abs/2007.09545, 2020.

\bibitem[Chang and Sun(2024)]{Chang2024Text2GraspGS}
Xiaoyun Chang and Yi Sun.
\newblock Text2grasp: Grasp synthesis by text prompts of object grasping parts.
\newblock \emph{ArXiv}, abs/2404.15189, 2024.

\bibitem[Chao et~al.(2021)Chao, Yang, Xiang, Molchanov, Handa, Tremblay, Narang, Wyk, Iqbal, Birchfield, Kautz, and Fox]{Chao2021DexYCBAB}
Yu-Wei Chao, Wei Yang, Yu Xiang, Pavlo Molchanov, Ankur Handa, Jonathan Tremblay, Yashraj~S. Narang, Karl~Van Wyk, Umar Iqbal, Stan Birchfield, Jan Kautz, and Dieter Fox.
\newblock Dexycb: A benchmark for capturing hand grasping of objects.
\newblock \emph{2021 IEEE/CVF Conference on Computer Vision and Pattern Recognition (CVPR)}, pages 9040--9049, 2021.

\bibitem[Dai et~al.(2018)Dai, Majumdar, and Tedrake]{Dai2018SynthsisOp}
Hongkai Dai, Anirudha Majumdar, and Russ Tedrake.
\newblock \emph{Synthesis and Optimization of Force Closure Grasps via Sequential Semidefinite Programming}, page 285–305.
\newblock 2018.

\bibitem[Duan et~al.(2021)Duan, Wang, Huang, Xu, Wei, and Shen]{Duan2021RoboticsDG}
Haonan Duan, Peng Wang, Yayu Huang, Guangyun Xu, Wei Wei, and Xiao Shen.
\newblock Robotics dexterous grasping: The methods based on point cloud and deep learning.
\newblock \emph{Frontiers in Neurorobotics}, 15, 2021.

\bibitem[Geng and Liu(2023)]{Geng2023UniDexGrasp++ID}
Haoran Geng and Yun Liu.
\newblock Unidexgrasp++: Improving dexterous grasping policy learning via geometry-aware curriculum and iterative generalist-specialist learning.
\newblock \emph{2023 IEEE/CVF International Conference on Computer Vision (ICCV)}, pages 3868--3879, 2023.

\bibitem[Geng et~al.(2022)Geng, Xu, Zhao, Xu, Yi, Huang, and Wang]{Geng2022GAPartNetCD}
Haoran Geng, Helin Xu, Chengyan Zhao, Chao Xu, Li Yi, Siyuan Huang, and He Wang.
\newblock Gapartnet: Cross-category domain-generalizable object perception and manipulation via generalizable and actionable parts.
\newblock \emph{2023 IEEE/CVF Conference on Computer Vision and Pattern Recognition (CVPR)}, pages 7081--7091, 2022.

\bibitem[Grady et~al.(2021)Grady, Tang, Twigg, Vo, Brahmbhatt, and Kemp]{Grady2021ContactOptOC}
Patrick Grady, Chengcheng Tang, Christopher~D. Twigg, Minh Vo, Samarth Brahmbhatt, and Charles~C. Kemp.
\newblock Contactopt: Optimizing contact to improve grasps.
\newblock \emph{2021 IEEE/CVF Conference on Computer Vision and Pattern Recognition (CVPR)}, pages 1471--1481, 2021.

\bibitem[Hampali et~al.(2019)Hampali, Rad, Oberweger, and Lepetit]{Hampali2019HOnnotateAM}
Shreyas Hampali, Mahdi Rad, Markus Oberweger, and Vincent Lepetit.
\newblock Honnotate: A method for 3d annotation of hand and object poses.
\newblock \emph{2020 IEEE/CVF Conference on Computer Vision and Pattern Recognition (CVPR)}, pages 3193--3203, 2019.

\bibitem[Hang et~al.(2024)Hang, Lin, Zhu, Li, Wu, Ma, and Sun]{hang2024dexfuncgrasp}
Jinglue Hang, Xiangbo Lin, Tianqiang Zhu, Xuanheng Li, Rina Wu, Xiaohong Ma, and Yi Sun.
\newblock Dexfuncgrasp: A robotic dexterous functional grasp dataset constructed from a cost-effective real-simulation annotation system.
\newblock In \emph{Proceedings of the AAAI Conference on Artificial Intelligence}, pages 10306--10313, 2024.

\bibitem[Hasson et~al.(2019)Hasson, Varol, Tzionas, Kalevatykh, Black, Laptev, and Schmid]{Hasson2019Obman}
Yana Hasson, G{\"u}l Varol, Dimitrios Tzionas, Igor Kalevatykh, Michael~J. Black, Ivan Laptev, and Cordelia Schmid.
\newblock Learning joint reconstruction of hands and manipulated objects.
\newblock \emph{2019 IEEE/CVF Conference on Computer Vision and Pattern Recognition (CVPR)}, pages 11799--11808, 2019.

\bibitem[Jian et~al.(2023)Jian, Liu, Li, Hu, and Liu]{Jian2023AffordPoseAL}
Juntao Jian, Xiuping Liu, Manyi Li, Ruizhen Hu, and Jian Liu.
\newblock Affordpose: A large-scale dataset of hand-object interactions with affordance-driven hand pose.
\newblock In \emph{Proceedings of the IEEE/CVF International Conference on Computer Vision (ICCV)}, pages 14713--14724, 2023.

\bibitem[Jiang et~al.(2021)Jiang, Liu, Wang, and Wang]{Jiang2021HandObjectCContactTTA}
Hanwen Jiang, Shaowei Liu, Jiashun Wang, and Xiaolong Wang.
\newblock Hand-object contact consistency reasoning for human grasps generation.
\newblock \emph{2021 IEEE/CVF International Conference on Computer Vision (ICCV)}, pages 11087--11096, 2021.

\bibitem[Karunratanakul et~al.(2020)Karunratanakul, Yang, Zhang, Black, Muandet, and Tang]{Karunratanakul2020GraspingField}
Korrawe Karunratanakul, Jinlong Yang, Yan Zhang, Michael~J. Black, Krikamol Muandet, and Siyu Tang.
\newblock Grasping field: Learning implicit representations for human grasps.
\newblock \emph{2020 International Conference on 3D Vision (3DV)}, pages 333--344, 2020.

\bibitem[Kenton and Toutanova(2019)]{kenton2019bert}
Jacob Devlin Ming-Wei~Chang Kenton and Lee~Kristina Toutanova.
\newblock Bert: Pre-training of deep bidirectional transformers for language understanding.
\newblock In \emph{Proceedings of naacL-HLT}, page~2. Minneapolis, Minnesota, 2019.

\bibitem[Li et~al.(2022{\natexlab{a}})Li, Lin, Zhou, Li, Huo, Chen, and Ye]{Li2022Contact2Grasp3G}
Haoming Li, Xinzhuo Lin, Yang Zhou, Xiang Li, Yuchi Huo, Jiming Chen, and Qi Ye.
\newblock Contact2grasp: 3d grasp synthesis via hand-object contact constraint.
\newblock In \emph{International Joint Conference on Artificial Intelligence}, 2022{\natexlab{a}}.

\bibitem[Li et~al.(2022{\natexlab{b}})Li, Baron, Zhang, and Rojas]{Li2022EfficientGraspAU}
Kelin Li, Nicholas Baron, Xianmin Zhang, and Nicol{\'a}s Rojas.
\newblock Efficientgrasp: A unified data-efficient learning to grasp method for multi-fingered robot hands.
\newblock \emph{IEEE Robotics and Automation Letters}, 7:\penalty0 8619--8626, 2022{\natexlab{b}}.

\bibitem[Li et~al.(2024)Li, Wang, Yang, Lu, and Dai]{Li2024SemGraspSG}
Kailin Li, Jingbo Wang, Lixin Yang, Cewu Lu, and Bo Dai.
\newblock Semgrasp: Semantic grasp generation via language aligned discretization.
\newblock \emph{ArXiv}, abs/2404.03590, 2024.

\bibitem[Li et~al.(2022{\natexlab{c}})Li, Zhang, Zhang, Yang, Li, Zhong, Wang, Yuan, Zhang, Hwang, et~al.]{li2022grounded}
Liunian~Harold Li, Pengchuan Zhang, Haotian Zhang, Jianwei Yang, Chunyuan Li, Yiwu Zhong, Lijuan Wang, Lu Yuan, Lei Zhang, Jenq-Neng Hwang, et~al.
\newblock Grounded language-image pre-training.
\newblock In \emph{Proceedings of the IEEE/CVF Conference on Computer Vision and Pattern Recognition}, pages 10965--10975, 2022{\natexlab{c}}.

\bibitem[Li et~al.(2022{\natexlab{d}})Li, Liu, Li, Geng, Zhu, Yang, and Huang]{Li2022GenDexGraspGD}
Puhao Li, Tengyu Liu, Yuyang Li, Yiran Geng, Yixin Zhu, Yaodong Yang, and Siyuan Huang.
\newblock Gendexgrasp: Generalizable dexterous grasping.
\newblock \emph{2023 IEEE International Conference on Robotics and Automation (ICRA)}, pages 8068--8074, 2022{\natexlab{d}}.

\bibitem[Liu et~al.(2023{\natexlab{a}})Liu, Li, Wu, and Lee]{Liu2023LLaVA}
Haotian Liu, Chunyuan Li, Qingyang Wu, and Yong~Jae Lee.
\newblock Visual instruction tuning.
\newblock \emph{ArXiv}, abs/2304.08485, 2023{\natexlab{a}}.

\bibitem[Liu et~al.(2019)Liu, Pan, Xu, Ganguly, and Manocha]{Liu2019GeneratingGP}
Min Liu, Zherong Pan, Kai Xu, Kanishka Ganguly, and Dinesh Manocha.
\newblock Generating grasp poses for a high-dof gripper using neural networks.
\newblock \emph{2019 IEEE/RSJ International Conference on Intelligent Robots and Systems (IROS)}, pages 1518--1525, 2019.

\bibitem[Liu et~al.(2023{\natexlab{b}})Liu, Zhu, Cai, Han, Ling, Porikli, and Su]{liu2023partslip}
Minghua Liu, Yinhao Zhu, Hong Cai, Shizhong Han, Zhan Ling, Fatih Porikli, and Hao Su.
\newblock Partslip: Low-shot part segmentation for 3d point clouds via pretrained image-language models.
\newblock In \emph{Proceedings of the IEEE/CVF conference on computer vision and pattern recognition}, pages 21736--21746, 2023{\natexlab{b}}.

\bibitem[Liu et~al.(2023{\natexlab{c}})Liu, Zhou, Yang, Gupta, and Wang]{Liu2023ContactGenGC}
Shaowei Liu, Yang Zhou, Jimei Yang, Saurabh Gupta, and Shenlong Wang.
\newblock Contactgen: Generative contact modeling for grasp generation.
\newblock \emph{2023 IEEE/CVF International Conference on Computer Vision (ICCV)}, pages 20552--20563, 2023{\natexlab{c}}.

\bibitem[Liu et~al.(2021)Liu, Liu, Jiao, Zhu, and Zhu]{Liu2021SynthesizingDA}
Tengyu Liu, Zeyu Liu, Ziyuan Jiao, Yixin Zhu, and Song-Chun Zhu.
\newblock Synthesizing diverse and physically stable grasps with arbitrary hand structures using differentiable force closure estimator.
\newblock \emph{IEEE Robotics and Automation Letters}, 7:\penalty0 470--477, 2021.

\bibitem[Lu et~al.(2023)Lu, Kang, Li, Liu, Yang, Huang, and Hua]{Lu2023UGGUG}
Jiaxin Lu, Hao Kang, Haoxiang Li, Bo Liu, Yiding Yang, Qixing Huang, and Gang Hua.
\newblock Ugg: Unified generative grasping.
\newblock \emph{ArXiv}, abs/2311.16917, 2023.

\bibitem[Miller and Allen(2004)]{Miller2004Graspit!}
Andrew~T. Miller and Peter~K. Allen.
\newblock Graspit! a versatile simulator for robotic grasping.
\newblock \emph{IEEE Robotics \& Automation Magazine}, 11:\penalty0 110--122, 2004.

\bibitem[Qi et~al.(2017)Qi, Yi, Su, and Guibas]{qi2017pointnet++}
Charles~Ruizhongtai Qi, Li Yi, Hao Su, and Leonidas~J Guibas.
\newblock Pointnet++: Deep hierarchical feature learning on point sets in a metric space.
\newblock \emph{Advances in neural information processing systems}, 30, 2017.

\bibitem[Radford et~al.(2021)Radford, Kim, Hallacy, Ramesh, Goh, Agarwal, Sastry, Askell, Mishkin, Clark, Krueger, and Sutskever]{Radford2021Clip}
Alec Radford, Jong~Wook Kim, Chris Hallacy, Aditya Ramesh, Gabriel Goh, Sandhini Agarwal, Girish Sastry, Amanda Askell, Pamela Mishkin, Jack Clark, Gretchen Krueger, and Ilya Sutskever.
\newblock Learning transferable visual models from natural language supervision.
\newblock In \emph{International Conference on Machine Learning}, 2021.

\bibitem[Rajeswaran et~al.(2017)Rajeswaran, Kumar, Gupta, Schulman, Todorov, and Levine]{Rajeswaran2017LearningCD}
Aravind Rajeswaran, Vikash Kumar, Abhishek Gupta, John Schulman, Emanuel Todorov, and Sergey Levine.
\newblock Learning complex dexterous manipulation with deep reinforcement learning and demonstrations.
\newblock \emph{ArXiv}, abs/1709.10087, 2017.

\bibitem[Romero and Tzionas(2017)]{Romero2017MANO}
Javier Romero and Dimitrios Tzionas.
\newblock Embodied hands : Modeling and capturing hands and bodies together * * supplementary material * *.
\newblock 2017.

\bibitem[She et~al.(2022)She, Hu, Xu, Liu, Xu, and Huang]{She2022LearningHR}
Qijin She, Ruizhen Hu, Juzhan Xu, Min Liu, Kai Xu, and Hui Huang.
\newblock Learning high-dof reaching-and-grasping via dynamic representation of gripper-object interaction.
\newblock \emph{ACM Transactions on Graphics (TOG)}, 41:\penalty0 1 -- 14, 2022.

\bibitem[Taheri et~al.(2020)Taheri, Ghorbani, Black, and Tzionas]{Taheri2020GRABAD}
Omid Taheri, Nima Ghorbani, Michael~J. Black, and Dimitrios Tzionas.
\newblock Grab: A dataset of whole-body human grasping of objects.
\newblock In \emph{European Conference on Computer Vision}, 2020.

\bibitem[Tang et~al.(2023)Tang, Huang, Ge, Liu, and Zhang]{Tang2023GraspGPTLS}
Chao Tang, Dehao Huang, Wenqiang Ge, Weiyu Liu, and Hong Zhang.
\newblock Graspgpt: Leveraging semantic knowledge from a large language model for task-oriented grasping.
\newblock \emph{IEEE Robotics and Automation Letters}, 8:\penalty0 7551--7558, 2023.

\bibitem[Tekden et~al.(2023)Tekden, Deisenroth, and Bekiroglu]{Tekden2023GraspTransferB}
Ahmet~E. Tekden, Marc~Peter Deisenroth, and Yasemin Bekiroglu.
\newblock Grasp transfer based on self-aligning implicit representations of local surfaces.
\newblock \emph{IEEE Robotics and Automation Letters}, 8:\penalty0 6315--6322, 2023.

\bibitem[Tse et~al.(2022)Tse, Zhang, Kim, Leonardis, Zheng, and Chang]{Tse2022S2ContactGN}
Tze Ho~Elden Tse, Zhongqun Zhang, Kwang~In Kim, Ale{\vs} Leonardis, Feng Zheng, and Hyung~Jin Chang.
\newblock S2contact: Graph-based network for 3d hand-object contact estimation with semi-supervised learning.
\newblock \emph{ArXiv}, abs/2208.00874, 2022.

\bibitem[Turpin et~al.(2022)Turpin, Wang, Heiden, Chen, Macklin, Tsogkas, Dickinson, and Garg]{Turpin2022GraspD}
Dylan Turpin, Liquan Wang, Eric Heiden, Yun-Chun Chen, Miles Macklin, Stavros Tsogkas, Sven Dickinson, and Animesh Garg.
\newblock Grasp’d: Differentiable contact-rich grasp synthesis for\&nbsp;multi-fingered hands.
\newblock In \emph{Computer Vision – ECCV 2022: 17th European Conference, Tel Aviv, Israel, October 23–27, 2022, Proceedings, Part VI}, page 201–221, Berlin, Heidelberg, 2022. Springer-Verlag.

\bibitem[Wang et~al.(2022)Wang, Zhang, Chen, Xu, Li, Liu, and Wang]{Wang2022DexGraspNetAL}
Ruicheng Wang, Jialiang Zhang, Jiayi Chen, Yinzhen Xu, Puhao Li, Tengyu Liu, and He Wang.
\newblock Dexgraspnet: A large-scale robotic dexterous grasp dataset for general objects based on simulation.
\newblock \emph{2023 IEEE International Conference on Robotics and Automation (ICRA)}, pages 11359--11366, 2022.

\bibitem[Wei et~al.(2022)Wei, Li, Wang, Li, Li, Luo, and Zhong]{Wei2022DVGGDV}
Wei Wei, Daheng Li, Peng Wang, Yiming Li, Wanyi Li, Yongkang Luo, and Jun Zhong.
\newblock Dvgg: Deep variational grasp generation for dextrous manipulation.
\newblock \emph{IEEE Robotics and Automation Letters}, 7:\penalty0 1659--1666, 2022.

\bibitem[Wei et~al.(2024)Wei, Jiang, Xing, Tan, Wu, Li, Cutkosky, and Zheng]{Wei2024DexGYGrasp}
Yi-Lin Wei, Jian-Jian Jiang, Chengyi Xing, Xiantuo Tan, Xiao-Ming Wu, Hao Li, Mark~R. Cutkosky, and Wei-Shi Zheng.
\newblock Grasp as you say: Language-guided dexterous grasp generation.
\newblock \emph{ArXiv}, abs/2405.19291, 2024.

\bibitem[Wei et~al.(2025)Wei, Lin, Lin, Jiang, Wu, Zeng, and Zheng]{wei2025afforddexgrasp}
Yi-Lin Wei, Mu Lin, Yuhao Lin, Jian-Jian Jiang, Xiao-Ming Wu, Ling-An Zeng, and Wei-Shi Zheng.
\newblock Afforddexgrasp: Open-set language-guided dexterous grasp with generalizable-instructive affordance.
\newblock \emph{arXiv preprint arXiv:2503.07360}, 2025.

\bibitem[Weng et~al.(2024)Weng, Lu, Kragic, and Lundell]{Weng2024DexDiffuserGD}
Zehang Weng, Haofei Lu, Danica Kragic, and Jens Lundell.
\newblock Dexdiffuser: Generating dexterous grasps with diffusion models.
\newblock \emph{ArXiv}, abs/2402.02989, 2024.

\bibitem[Wohlkinger et~al.(2012)Wohlkinger, Aldoma, Rusu, and Vincze]{Wohlkinger20123DNetLO}
Walter Wohlkinger, Aitor Aldoma, Radu~Bogdan Rusu, and Markus Vincze.
\newblock 3dnet: Large-scale object class recognition from cad models.
\newblock \emph{2012 IEEE International Conference on Robotics and Automation}, pages 5384--5391, 2012.

\bibitem[Xiang et~al.(2020)Xiang, Qin, Mo, Xia, Zhu, Liu, Liu, Jiang, Yuan, Wang, Yi, Chang, Guibas, and Su]{Xiang2020SAPIEN}
Fanbo Xiang, Yuzhe Qin, Kaichun Mo, Yikuan Xia, Hao Zhu, Fangchen Liu, Minghua Liu, Hanxiao Jiang, Yifu Yuan, He Wang, Li Yi, Angel~X. Chang, Leonidas~J. Guibas, and Hao Su.
\newblock {SAPIEN}: A simulated part-based interactive environment.
\newblock In \emph{The IEEE Conference on Computer Vision and Pattern Recognition (CVPR)}, 2020.

\bibitem[Xu et~al.(2024)Xu, Wei, Zheng, Wu, and Zheng]{xu2024dexterous}
Guo-Hao Xu, Yi-Lin Wei, Dian Zheng, Xiao-Ming Wu, and Wei-Shi Zheng.
\newblock Dexterous grasp transformer.
\newblock In \emph{Proceedings of the IEEE/CVF Conference on Computer Vision and Pattern Recognition}, pages 17933--17942, 2024.

\bibitem[Xu et~al.(2023)Xu, Wan, Zhang, Liu, Shan, Shen, Wang, Geng, Weng, Chen, Liu, Yi, and Wang]{Xu2023UniDexGraspUR}
Yinzhen Xu, Weikang Wan, Jialiang Zhang, Haoran Liu, Zikang Shan, Hao Shen, Ruicheng Wang, Haoran Geng, Yijia Weng, Jiayi Chen, Tengyu Liu, Li Yi, and He Wang.
\newblock Unidexgrasp: Universal robotic dexterous grasping via learning diverse proposal generation and goal-conditioned policy.
\newblock \emph{2023 IEEE/CVF Conference on Computer Vision and Pattern Recognition (CVPR)}, pages 4737--4746, 2023.

\bibitem[Yang et~al.(2020)Yang, Zhan, Li, Xu, Li, and Lu]{Yang2020CPFLA}
Lixin Yang, Xinyu Zhan, Kailin Li, Wenqiang Xu, Jiefeng Li, and Cewu Lu.
\newblock Cpf: Learning a contact potential field to model the hand-object interaction.
\newblock \emph{2021 IEEE/CVF International Conference on Computer Vision (ICCV)}, pages 11077--11086, 2020.

\bibitem[Yang et~al.(2022)Yang, Li, Zhan, Wu, Xu, Liu, and Lu]{Yang2022OakInkAL}
Lixin Yang, Kailin Li, Xinyu Zhan, Fei Wu, Anran Xu, Liu Liu, and Cewu Lu.
\newblock Oakink: A large-scale knowledge repository for understanding hand-object interaction.
\newblock \emph{2022 IEEE/CVF Conference on Computer Vision and Pattern Recognition (CVPR)}, pages 20921--20930, 2022.

\bibitem[Zhang et~al.(2023)Zhang, Hang, Zhu, Lin, Wu, Peng, Tian, and Sun]{Zhang2023FunctionalGraspLF}
Yibiao Zhang, Jinglue Hang, Tianqiang Zhu, Xiangbo Lin, Rina Wu, Wanli Peng, Dongying Tian, and Yi Sun.
\newblock Functionalgrasp: Learning functional grasp for robots via semantic hand-object representation.
\newblock \emph{IEEE Robotics and Automation Letters}, 8:\penalty0 3094--3101, 2023.

\bibitem[Zhou et~al.(2023)Zhou, Gu, Li, Liu, Fang, and Su]{zhou2023partslip++}
Yuchen Zhou, Jiayuan Gu, Xuanlin Li, Minghua Liu, Yunhao Fang, and Hao Su.
\newblock Partslip++: Enhancing low-shot 3d part segmentation via multi-view instance segmentation and maximum likelihood estimation.
\newblock \emph{arXiv preprint arXiv:2312.03015}, 2023.

\bibitem[Zhu et~al.(2021)Zhu, Wu, Lin, and Sun]{Zhu2021TowardHG}
Tianqiang Zhu, Rina Wu, Xiangbo Lin, and Yi Sun.
\newblock Toward human-like grasp: Dexterous grasping via semantic representation of object-hand.
\newblock \emph{2021 IEEE/CVF International Conference on Computer Vision (ICCV)}, pages 15721--15731, 2021.

\end{thebibliography}
}

\end{document}

\clearpage
\setcounter{page}{1}
\maketitlesupplementary

% ---------------------------------------------------------------------------------------------------------------------------
\section{Prompt of MLLM}
\begin{figure}[!h]
   \includegraphics[width=1.05\linewidth]{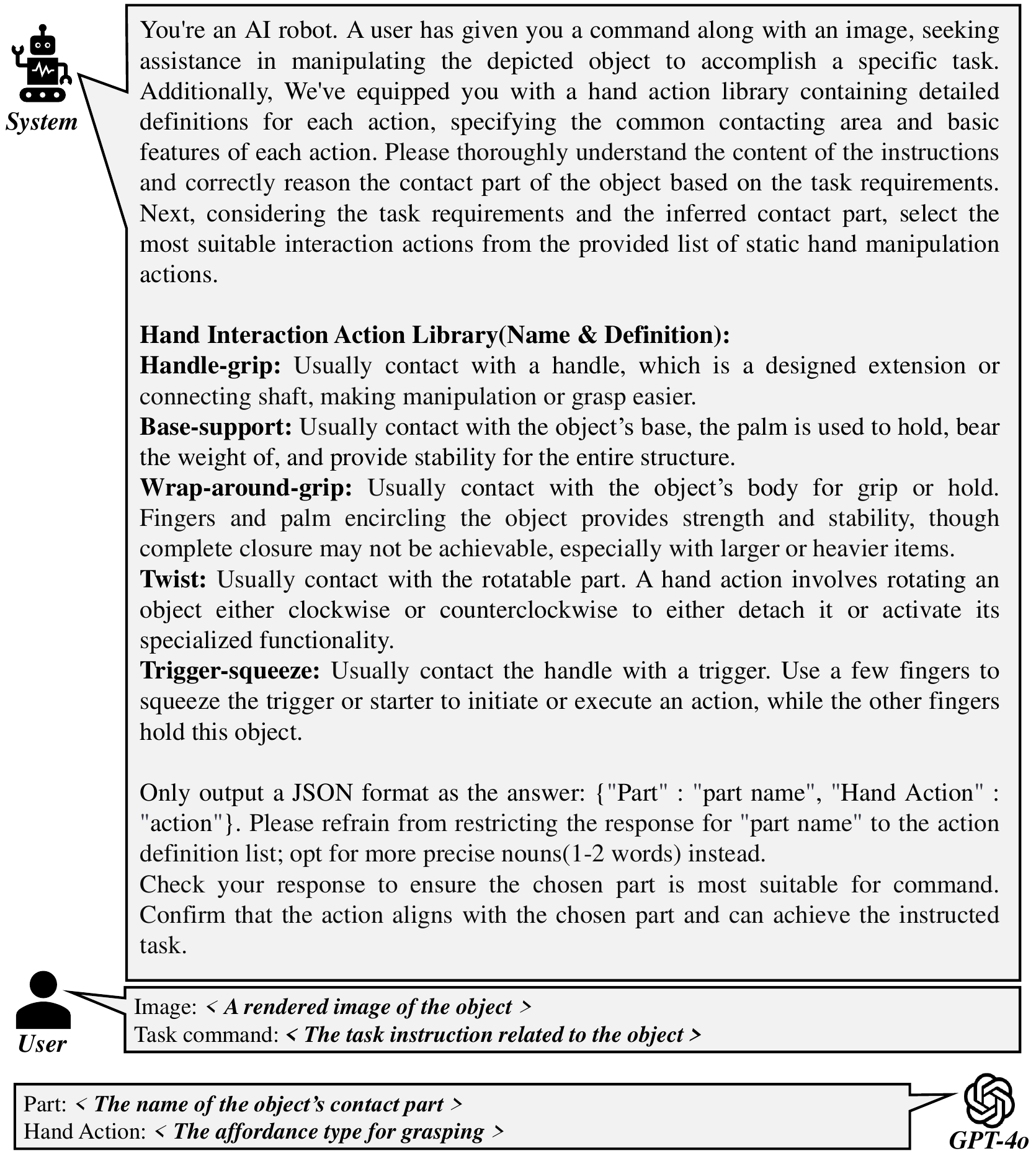}
   \caption{Prompt of MLLM for task instruction analysis. }
   \label{fig: supplementary-prompt}
\end{figure}

As mentioned in Section 3.2 of our main paper, we prompt the pre-trained MLLM to determine the fine-grained grasping arrangement, i.e. affordance type $A$ and the name of contact part $P$. We employ GPT-4o from OpenAI API for the task instruction analysis. The prompt we use is shown in Figure~\ref{fig: supplementary-prompt}.  By carefully describing the analysis purpose and the definition of the pre-defined action library, we can leverage the pre-trained MLLM to analyze the user-specified input images and instructions in a zero-shot manner, and identify the interaction parts as well as the appropriate action type. This setup relies only on the powerful commonsense reasoning capabilities of MLLM, without the need for human-provided examples.

% ----------------------------------------------------------------------------------------------------------------------------
\section{Perceptual Evaluation}
As mentioned in Section 4.1 of our main paper, we assess the perceptual score (Percep.Score $\%$) as one evaluation metric for which we conduct a user study of 20 groups of results with 30 participants to judge the quality and task alignment of generated grasps. The participants invited to rate based on the key aspects of \textbf{task semantic alignment}, \textbf{naturalness}, \textbf{plausibility}, and \textbf{human-likeness} of the synthesized hand configurations. Specifically, we randomly selected 20 experimental results across 19 distinct object categories that are unseen in the training stage for visualization.

\begin{figure}[!h]
   \includegraphics[width=0.9\linewidth]{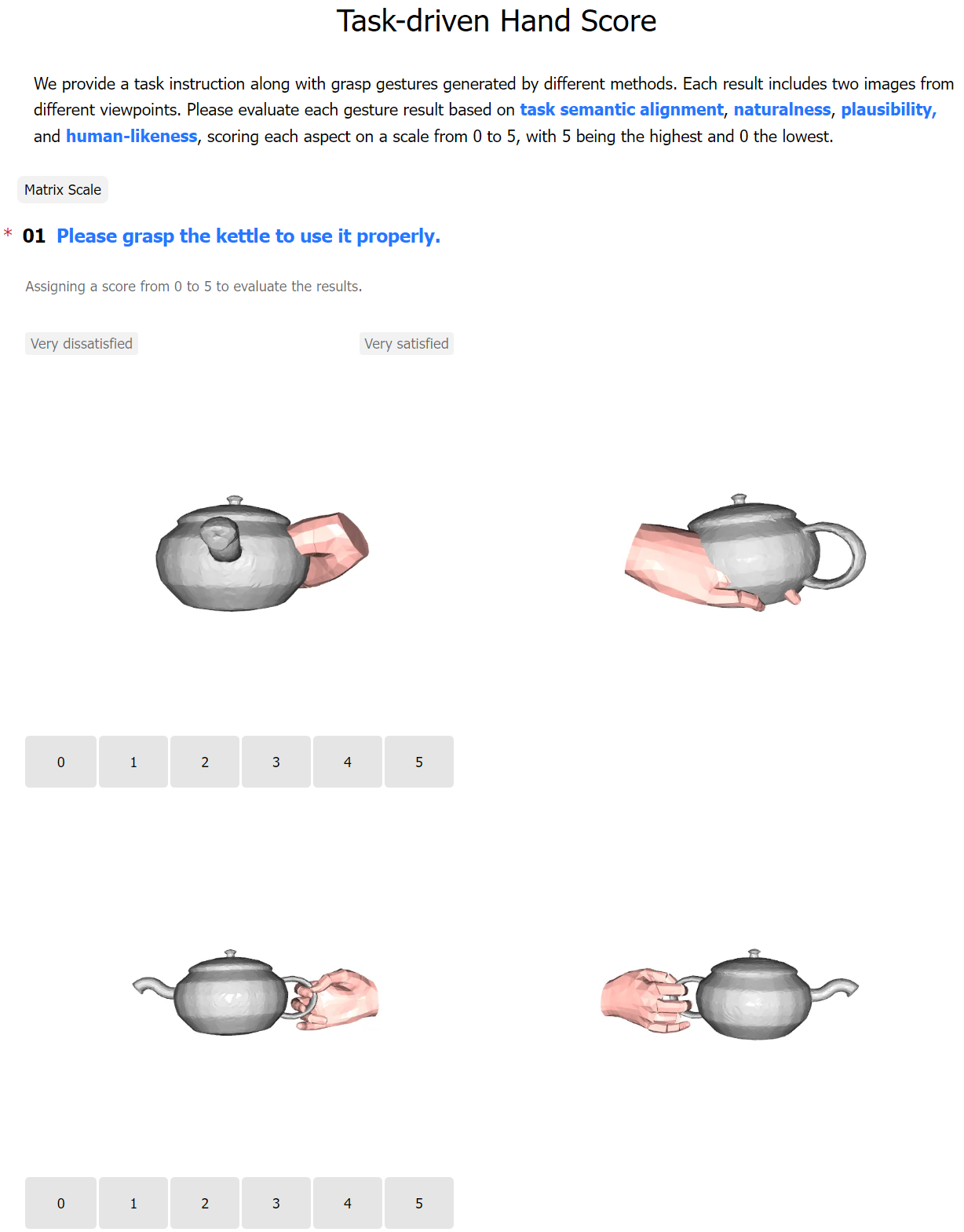}
   \caption{The partial screenshot of the user survey questionnaire. }
   \label{fig: supplementary-user study}
\end{figure}

For each object, we provided the specified textual instruction (e.g. \textit{``Please grasp the kettle to use it properly"}) and different experimental results: 5 from comparison experiments and 6 from ablation studies. Each result is rated independently. To facilitate observation, we rendered two images from different viewpoints for each experimental result. Volunteers were asked to rate each result on a scale of 0 to 5, where 5 indicates the highest score and 0 is the lowest. Fig.~\ref{fig: supplementary-user study} is a partial screenshot of the user survey questionnaire, which shows two different results for the kettle.

% -----------------------------------------------------------------------------------------------------------------------------
\section{More Results}
% \input{Supplementary/tables/supple_comparison}

% ----------------------------------------------------------------------------------------------------------------------------
\subsection{Intermediate Results}
\begin{figure*}[!t]
   \centering
   \includegraphics[width=0.95\linewidth]{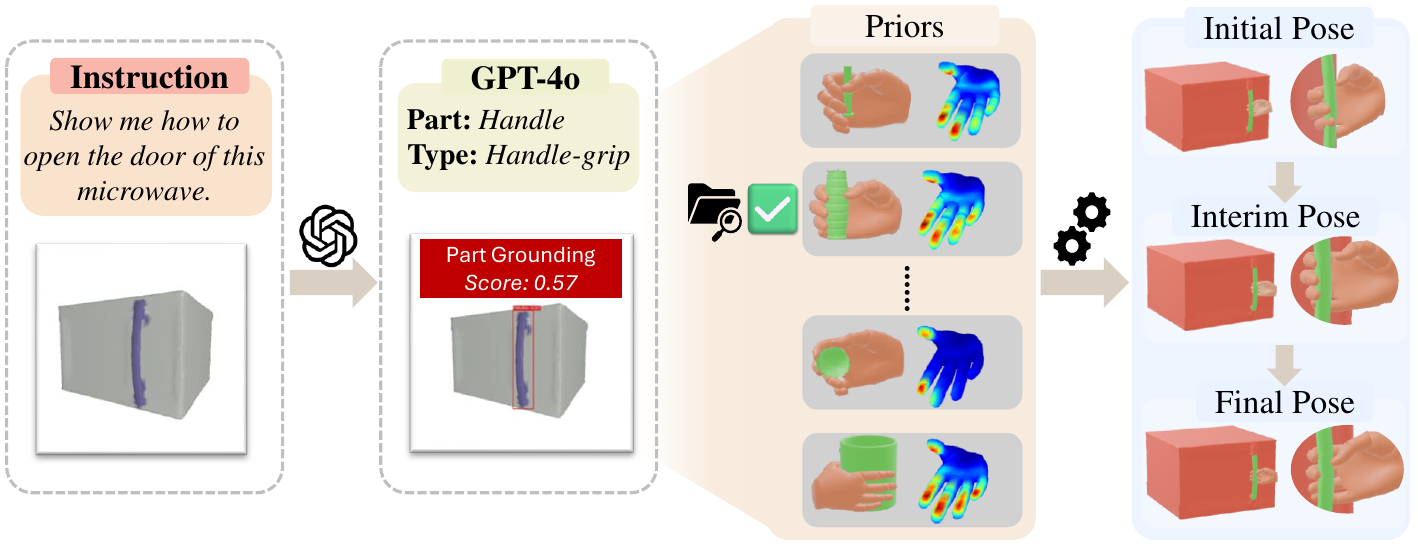}
   \caption{The intermediate results of our G-DexGrasp. (1) Model input (2) MLLM output and grounding results; (3) Grasp priors (including intrinsic params and hand contact maps) visualization (4) Predicted grasp poses in three stages}
   \label{fig: supplementary-intermediate}
\end{figure*}

To better show how G-DexGrasp worked, we take the results of a microwave as an example in Figure~\ref{fig: supplementary-intermediate}. First, we randomly color the parts of the object to improve visual perception for MLLM processing. Quantitative evaluations on unseen datasets using GPT-4o with our customized prompts achieve $96.99\%$ accuracy in grasp-type parsing and $76.6\%$ precision in combined part parsing and grounding. Subsequently, grasp priors associated with \emph{Handle-grip} and \emph{Handle} are accurately retrieved (OBB-Based). Notably, the grasp priors within a part of clusters have been showing, revealing significant variations strongly correlated with part semantics and tasks. Then, the Part-Aware Grasping Generation Network subsequently predicts intrinsic parameters, where initial poses achieve semantic-task alignment but exhibit penetration artifacts (e.g., pinky finger) and incomplete contacts (e.g., index finger). The first optimization stage refines extrinsic parameters with the grasp priors constraints, improving surface conformity while some penetrations and incomplete contacts remain. The second stage jointly adjusts intrinsic and extrinsic parameters, achieving a physically plausible, contact-compliant grasp.

Moreover, although the five affordance types already cover common grasp-related tasks and help avoid local optima during the optimization stage, we further introduce two additional hand-related affordance types, namely \emph{point} and \emph{press}. With these seven types, the affordance prediction accuracy reaches $96.66\%$ and the part grounding accuracy $75.25\%$, both comparable to the original five-type setting.

\begin{table}
\centering
% \begin{tabular}{l|ccc|cc}
\resizebox{0.48\textwidth}{!}{
\begin{tabular}{l|p{2.4cm}<{\centering}p{1.6cm}<{\centering}p{1.6cm}<{\centering}|p{1.5cm}<{\centering}}
\hline
\multirow{2}{*}{Experiments} & \multicolumn{3}{c|}{Quality and Stability}                                                                         & \multicolumn{1}{c}{Semantic} \\ 
\cline{2-5}
                             % & \makecell[c]{Solid.Intsec.Vol\\($cm^3$)$\downarrow$}   & \makecell[c]{Penet.Depth\\($cm$)$\downarrow$}        & \makecell[c]{Sim.Dis\\($cm$) $\downarrow$}            & \makecell[c]{Part Acc.\\$(\%)$} \\
                              & Solid.Intsec.Vol$\downarrow$   & Penet.Depth$\downarrow$        & Sim.Dis$\downarrow$      & Part Acc. \\ 
\hline
Ours-OBB                                           & 2.94               & 0.29               & 4.27               & 71.6     \\
Ours-BPS                                            & 2.97              & 0.38            & 4.53                     & 71.6                \\
\hline
\end{tabular}
}
\caption{Quantitative comparison. Ours uses pre-trained models to infer and localize the contact parts, while Our-OBB and Our-BPS refer to representing part with OBB size and encoded features using BPS, respectively.}
\label{tab: bps-based results}
\end{table}
\subsection{BPS-based Results}
In Section 3.2 of the main paper, we employ the part-level OBB size for dataset clustering and prior retrieval, which is simple yet effective as shown in the results. Notably, it is easily replaced with more advanced geometric descriptors. For example, we implemented the Basis Point Set (BPS)~\cite{Prokudin2019EfficientBPSLO} to encode 3D part shapes, and measuring inter-part feature distances using cosine similarity. As shown in Table~\ref{tab: bps-based results}, the experimental results show that this BPS-based approach performs similarly to the OBB-based method while significantly outperforming other comparative approaches (shown in the main paper) across multiple evaluation metrics.

% (shown in Table~\ref{tab: supple_comparison_results}) across multiple evaluation metrics.

% ----------------------------------------------------------------------------------------------------------------------------
\begin{table}[!t]
\centering
\footnotesize
% \scalebox{1.0}{
\resizebox{0.40\textwidth}{!}{
\begin{tabular}{@{}l|cccc@{}}
\hline
Diversity    & $T_{std}$ ({$\uparrow$})    & $R_{std}$ ({$\uparrow$})      & $J_{std}$ ({$\uparrow$})    \\ 
\hline
AffordPose    & 0.49              & 40.91            & 12.94               \\ 
GrabNet$^\dag$    & 0.47          & 37.51         & 13.21               \\ 
GraspTTA$^\dag$     & 0.43        & 65.51          & 17.43            \\ 
\hline
Ours    & 0.57    & 78.57   & 13.18        \\ 
\hline
\end{tabular}
}
\caption{Quantitative diversity metrics compared to the baseline based on the test-set of AffordPose dataset. $T_{std}$, $R_{std}$ and $J_{std}$ represent the standard deviations of translation, rotation, and joint angles, respectively.}
\label{tab:diversity-tab}
\end{table}
\subsection{Diversity Results}
Our work focuses on task-driven semantic grasping, aiming to synthesize anthropic and reasonable hand grasps that facilitate task completion. While we do not explicitly pursue grasp diversity - as excessive variation may result in unnatural hand configurations misaligned with human preferences or compromised task performance - we quantify the diversity of synthesized hands through the standard deviations analysis of translation, rotation, and intrinsic joint angle. As demonstrated in Table~\ref{tab:diversity-tab}, our method exhibits competitive performance in translational and rotational diversity, while showing potential for improvement in joint angle variations. Notably, since baseline methods exhibit degraded performance on unseen test data, our diversity metrics are evaluated exclusively on the AffordPose test-set to ensure valid grasp references for comparative analysis.

% ----------------------------------------------------------------------------------------------------------------------------
% \begin{table}[!t]
% % \vspace{-5pt}
% % \vspace{-5pt}
% \centering
% \footnotesize
% % \scalebox{1.0}{
% \resizebox{0.48 \textwidth}{!}{
% \begin{tabular}{@{}l|ccc|c@{}}
% \toprule
% Diversity    & Solid.Intsec.Vol($cm^3$)$\downarrow$    & Penet.Depth($cm$)$\downarrow$      & Sim.Dis($cm$)$\downarrow$   & Part Acc.$(\%)$     \\ \midrule \midrule
% Objaverse-PartField    & 5.93    & 0.90   & 5.31       & 70.0   \\ 
% Few Intrinsics    & 2.35              & 0.27            & 5.91                & 70.23            \\ 
% Few Extrinsics    & 4.25             & 0.43            & 5.28                & 71.24            \\ 
% Fast Version     & 3.32    & 0.37   & 5.78       & 73.91   \\ \midrule
% Ours    & 2.94    & 0.29   & 4.27       & 71.6   \\ \bottomrule
% \end{tabular}
% }
% \caption{More Quantitative evaluation results.}
% \label{tab:rebuttal-table}
% \vspace{-10pt}
% \end{table}

\begin{table}
\centering
% \begin{tabular}{l|ccc|cc}
\resizebox{0.48\textwidth}{!}{
\begin{tabular}{l|p{2.4cm}<{\centering}p{1.6cm}<{\centering}p{1.6cm}<{\centering}|p{1.5cm}<{\centering}}
\hline
\multirow{2}{*}{Experiments} & \multicolumn{3}{c|}{Quality and Stability}                                                                         & \multicolumn{1}{c}{Semantic} \\ 
\cline{2-5}
                             % & \makecell[c]{Solid.Intsec.Vol\\($cm^3$)$\downarrow$}   & \makecell[c]{Penet.Depth\\($cm$)$\downarrow$}        & \makecell[c]{Sim.Dis\\($cm$) $\downarrow$}            & \makecell[c]{Part Acc.\\$(\%)$} \\
                              & Solid.Intsec.Vol$\downarrow$   & Penet.Depth$\downarrow$        & Sim.Dis$\downarrow$      & Part Acc. \\ 
\hline
Objaverse-PartField    & 5.93    & 0.90   & 5.31       & 70.0   \\ 
Few Intrinsics    & 2.35              & 0.27            & 5.91                & 70.23            \\ 
Few Extrinsics    & 4.25             & 0.43            & 5.28                & 71.24            \\ 
Fast Version     & 3.32    & 0.37   & 5.78       & 73.91   \\ \hline
Ours    & 2.94    & 0.29   & 4.27       & 71.6   \\
\hline
\end{tabular}
}
\caption{Quantitative evaluation for generalization and scalability. Objaverse-PartField denotes the full model evaluated on the open-set Objaverse dataset and pre-segmented by PartField. “Few-Intrinsics” and “Few-Extrinsics” refer to reduced supervision settings with limited intrinsic and extrinsic parameters, respectively.}
\label{tab:rebuttal-table}
\end{table}

\subsection{Evaluation on Open-set Dataset}
To better evaluate generalization in open-set scenarios, we select the first 500 objects from the Objaverse-tiny dataset~\cite{deitke2023objaverse} and filter out semantically ambiguous or non-graspable instances (e.g., boats). The remaining objects are resized to a canonical scale and pre-segmented using PartField~\cite{liu2025partfield} and subsequently processed through our proposed pipeline. The resulting performance on this setting, referred to as \emph{Objaverse-PartField}, is comparable to that on our unseen test set (see Table~\ref{tab:rebuttal-table}), highlighting the robustness and generalizability of our approach.

% ----------------------------------------------------------------------------------------------------------------------------
\subsection{Evaluations for Scalability}
We conduct the following experiments to validate the scalability of our method:

\noindent\emph{``Few-Intrinsics"}: We reserve only 30 intrinsic parameters per affordance type for the retrieval, which is approximately 1/100 of our original dataset.

\noindent\emph{``Few-Extrinsics"}: We use only 1/10 of the extrinsic parameters from the original dataset to train the network.

These experiments perform comparably with our full approach and much better than the baselines in our main paper, validating that our approach is scalable with a small set of fine-grained labels. However, once the retrieved prior is completely removed, the performance decreases dramatically as the experiment "w/o prior guide." in the main paper.

% ----------------------------------------------------------------------------------------------------------------------------
\subsection{More Visual Results}
\begin{figure*}[!t]
   \includegraphics[width=1.0\linewidth]{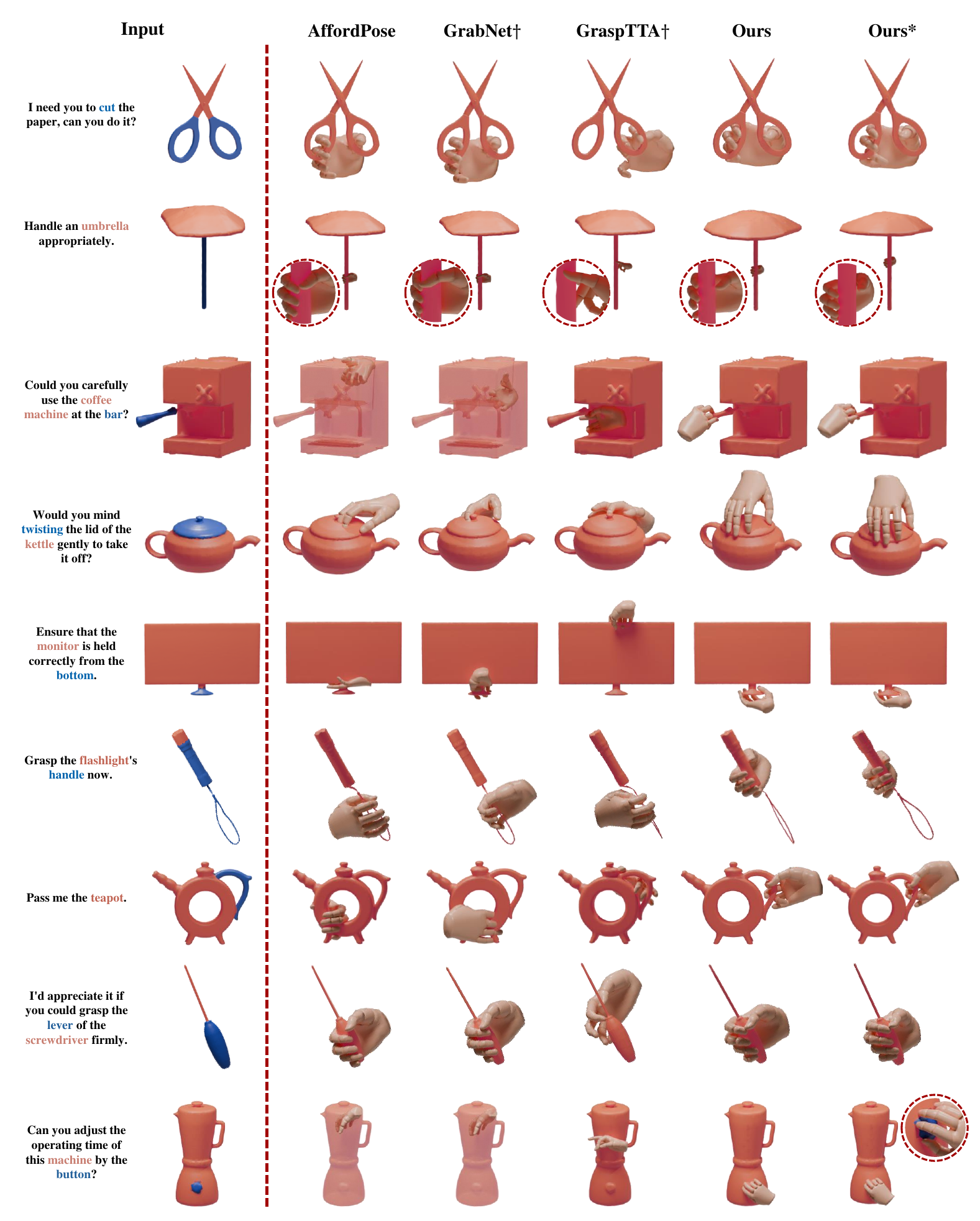}
   \caption{More visual results of comparison experiments. The parts that align with the task instructions are highlighted in blue.}
   \label{fig: supplementary-comparison}
\end{figure*}

Figure~\ref{fig: supplementary-comparison} presents comparison results on additional object categories. The first row shows grasps on a scissors from a seen category, where all methods generate reasonable poses and contact regions, except GraspTTA~\cite{Jiang2021HandObjectCContactTTA}, which produces an unstable grasp. The remaining rows show unseen categories, where existing methods often yield unreasonable results with severe interpenetration and grasp failures when test data deviates from training data.
% shows the results of the comparison experiments on more object categories. The first row visualizes the generated grasps on a scissors, which comes from a seen category during training. For this case, all the generated results exhibit reasonable hand poses and contact regions, except that GraspTTA~\cite{Jiang2021HandObjectCContactTTA} makes an unstable grasp. The rest rows visualize results on more unseen object categories. It is obvious that the existing approaches often generate unreasonable results, causing severe object interpenetration and grasp failures when the test data is not close to the data in the training dataset.

\begin{figure*}[!t]
   \includegraphics[width=1.0\linewidth]{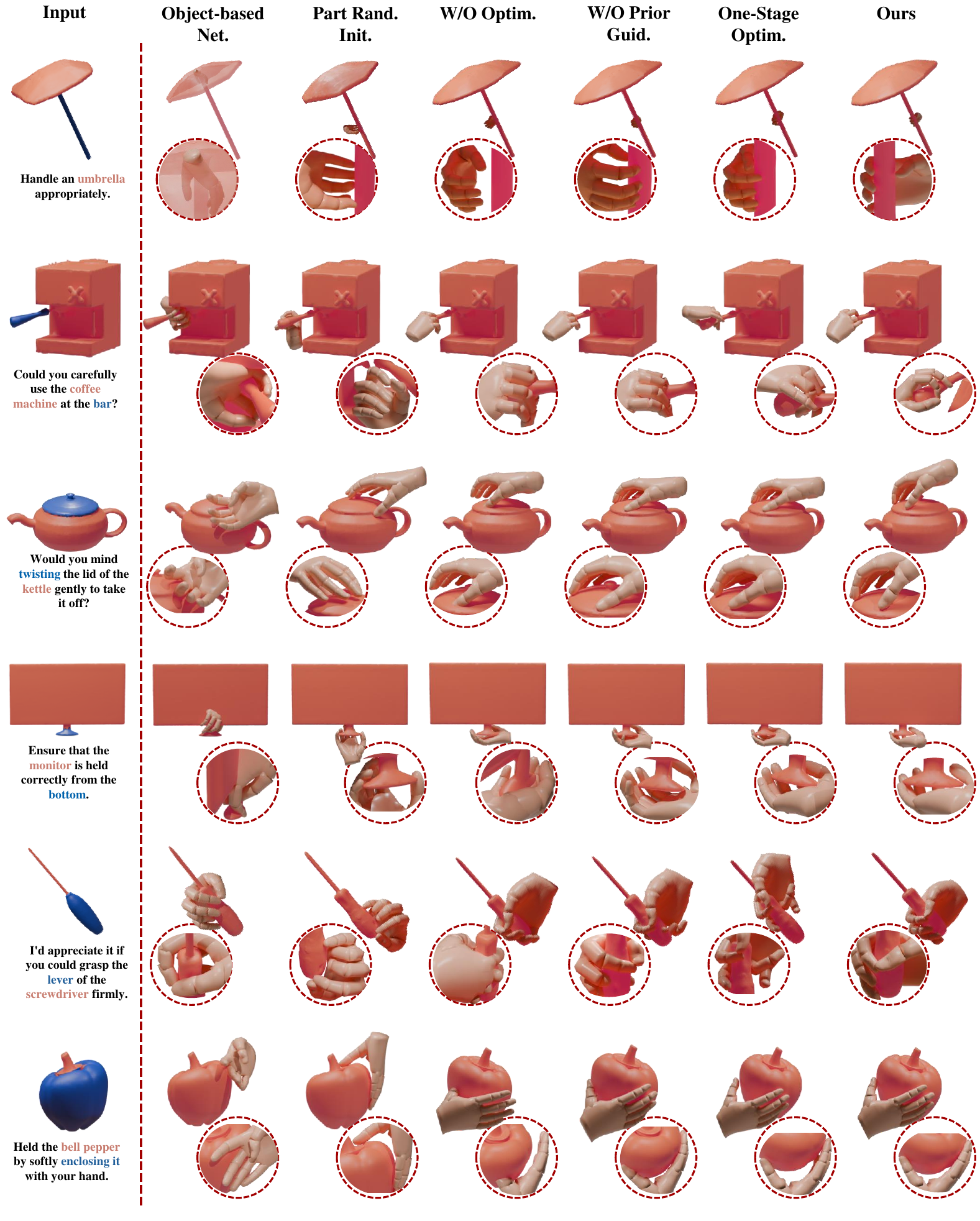}
   \caption{More visual results of ablation experiments. The parts that align with the task instructions are highlighted in blue. The dashed circles in the lower corner of each result show the corresponding partial enlarged views.}
   \label{fig: supplementary-ablation}
\end{figure*}

Figure~\ref{fig: supplementary-ablation} shows the ablation experiment results on the same set of objects from unseen categories. The first two experiments, i.e. Object-based Net. and Part Rand Init., produce obviously worse results. Once the dexterous hands are absurdly initialized around the object, it is difficult to optimize the extrinsic and intrinsic parameters to correct them, such as the umbrella case in the first row. By contrast, without a refinement optimization, i.e. W/O Optim., although the generated grasps look reasonable, it is noticeable that these grasp cannot stably hold the objects for the subsequent tasks, see the umbrella and kettle cases, while it remains severe penetration, see the bell pepper, coffee-machine cases. The last two experiments, W/O Prior Guid. and One-Stage Optim., produces nearly satisfactory results as ours. However, from the zoom-in visualizations, we can see that these generated results are actually un-optimal, also with remaining penetration (e.g., the W/O Prior Guid. results in the second and the fifth row), unstable hand intrinsics (e.g., the W/O Prior Guid. and the One-Stage Optim. result for the umbrella case and the screwdriver case in the fifth row), and discontaction between hand and object (e.g., the One-Stage Optim. results for bell pepper in the last row), etc.


{
    \small
    \bibliographystyle{ieeenat_fullname}
    \bibliography{X_suppl}
}
